\documentclass[preprint]{elsarticle}

\usepackage{graphicx}
\usepackage{amssymb}
\usepackage{amsmath}
\usepackage{url}
\usepackage{tabularx}
\usepackage{subfig}
\usepackage{booktabs} 
\usepackage[numbers]{natbib}
\usepackage{xcolor}

\usepackage{lineno,hyperref}

\newcommand{\etal}{\textit{et al.}}

\begin{document}

\begin{frontmatter}

\title{The SARAS Endoscopic Surgeon Action Detection (ESAD) dataset: Challenges and methods}



\author[label1]{Vivek Singh Bawa}
\ead{vsingh@brookes.ac.uk}
\author[label2]{Gurkirt Singh}
\author[label1]{Francis Kaping'A}
\author[label1]{Inna Skarga-Bandurova}
\author[label4]{Elettra Oleari}
\author[label4]{Alice Leporini}
\author[label4]{Carmela Landolfo}
\author[label6]{Pengfei Zhao}
\author[label6]{Xi Xiang}
\author[label6]{Gongning Luo}
\author[label6]{Kuanquan Wang}
\author[label7]{Liangzhi Li}
\author[label7]{Bowen Wang}
\author[label8]{Shang Zhao}
\author[label8]{Li Li}

\author[label4]{Armando Stabile}
\author[label5]{Francesco Setti}
\author[label5]{Riccardo Muradore}
\author[label1]{Fabio Cuzzolin}

\address[label1]{Visual Artificial Intelligence Laboratory, Oxford Brookes University, UK}
\address[label2]{Computer Vision Lab, ETH Zurich, Switzerland}
\address[label4]{San Raffaele Hospital, Milan, Italy}
\address[label5]{University of Verona, Italy}
\address[label6]{Harbin Institute of Technology, China}
\address[label8]{The George Washington University, USA}
\address[label7]{Osaka University, Japan}


\begin{abstract}
For an autonomous robotic system, monitoring surgeon actions and assisting the main surgeon during a procedure can be very challenging. The challenges come from the peculiar structure of the surgical scene, the greater similarity in appearance of actions performed via tools in a cavity compared to, say, human actions in unconstrained environments, as well as from the motion of the endoscopic camera. 
This paper presents ESAD, the first large-scale dataset designed to tackle the problem of surgeon action detection in endoscopic minimally invasive surgery. 
ESAD aims at contributing to increase the effectiveness and reliability of surgical assistant robots by realistically testing their awareness of the actions performed by a surgeon. The dataset provides bounding box annotation for 21 action classes on real endoscopic video frames captured during prostatectomy, and was used as the basis of a recent MIDL 2020 challenge. 
We also present an analysis of the dataset conducted using the baseline model which was released as part of the challenge, and a description of the top performing models submitted to the challenge together with the results they obtained. This study provides significant insight into what approaches can be effective and can be extended further. We believe that ESAD will serve in the future as a useful benchmark for all researchers active in surgeon action detection and assistive robotics at large.
\end{abstract}

\begin{keyword}
Surgeon action detection \sep endoscopic video \sep deep learning \sep prostatectomy \sep autonomy \sep computer vision \sep surgical robotics
\end{keyword}

\end{frontmatter}


\section{Introduction}
\label{S:1}


Minimally Invasive Surgery (MIS) is a very sensitive medical procedure, 
typically involving a main surgeon and an assistant surgeon. The success of an MIS procedure rests upon multiple factors, such as the attentiveness of the two surgeons, 
their competence, their degree of 
coordination, 
and so on. 
According to the Lancet Commission, each year 4.2 million people die within 30 days of surgery \cite{nepogodiev2019global}. Another study at Johns Hopkins University points out that 10\% of total deaths in the USA are due to medical error \cite{jhustudy}. There is no definite measure to compute and predict the risk factors associated with surgeon behaviour, making it very critical to monitor the set of actions performed by surgeons during a procedure in real time, so that any unfortunate event may be avoided.
Artificial intelligence is widely employed in applications where human error needs to be mitigated. The use of artificial intelligence in diagnostic imaging and electrodiagnosis, for instance, has been steadily rising in the past few years with the aim of mitigating human error issues \cite{jiang2017artificial, liew2018future}. There are growing demands to explore the application of artificial intelligence to a variety of other areas, such as healthcare delivery, healthcare administration, clinical decision support, patient monitoring and healthcare interventions, among others. \cite{reddy2019artificial, hughes2020artificial}. 

\emph{Robotic Minimally Invasive Surgery} (R-MIS), in particular, has surgical procedures performed remotely with the help of robotic arms using minimal surgical incisions to reduce trauma to the body. An R-MIS procedure involves at least five people (one main surgeon, one assistant surgeon, two nurses and one anaesthetist). Such an operation requires intensive communication among all these actors, highlighting the risk of error or miscommunication. 
The introduction of an autonomous robotic assistant surgeon \cite{leporini2020technical}, such as the one developed by the EU funded SARAS project (\url{https://saras-project.eu/}), has the potential to make surgical procedures safer. To accomplish that, the robotic assistant needs to identify and track the actions performed by the main surgeon in the surgical cavity and captured by the endoscopic camera. 

\emph{Action detection} is an establish field in computer vision. The problem
combines two tasks: (i) recognising an action and (ii) locating the action, which are tackled jointly. 
Action recognition consists in understanding what class of action is being performed, whereas action localisation provides the location of that action, typically on the image plane and in the form of a rectangular bounding box containing the action instance at hand. This gives complete information about what is happening and where. 
This knowledge can then be used in different ways, for example to drive the autonomous 
intervention of the robot during a surgery, to enable semi-supervised decision support and assistive robotics, to record and offline analyse the variations across different surgeries by computing useful statistics.



Various datasets exist to validate the action detection task in computer vision. However, they all focus on full-body actions performed by human beings captured by external cameras, and consider action classes that are fairly clearly distinguishable based on appearance 
(see for instance AVA \cite{gu2018ava}, DALY \cite{weinzaepfel2016human} or UCF101-24 \cite{soomro2012ucf101}). No dataset, however, has yet been devised to allow the validation of action detection in the medical domain, specifically for MIS or R-MIS surgery. 
\\
Action detection from endoscopic video exhibits a number of specific features.
Firstly, actions are not performed by humans directly but only through tele- or manually-operated tools. Secondly, the surgical cavity as a 'scene' is rather indistinct compared to the external scenes captured by classical datasets, in which objects of very different guise and shape can be easily discerned.
In an endoscopic video, instead, different organs 
will typically appear rather similar, with no clear boundaries, while surgical tools themselves may only be distinguished by tiny, subtle details of the jaws. Thirdly, endoscopic videos are captured at very close range, which 
prevents from obtaining sufficient contextual information for action recognition.

For all these reasons, surgeon action detection from endoscopic video is bound to be much more challenging than classical human action detection. Nevertheless, this hypothesis was never validated up to now, nor was a benchmark specifically designed to support the introduction of action detection in surgical robotics and medical imaging available.

\subsection{The ESAD dataset}

The ESAD dataset is the first benchmark explicitly designed to assess and evaluate methods for the detection of surgeon actions from endoscopic videos, developed with the assistance of medical professionals as well as expert surgeons.
The dataset contains four complete radical prostatectomy (RARP) procedures, each around 4 hours long, annotated with 46,325 action instances in the form of a bounding box with the associated action label. The dataset contemplates 21 action classes specific to radical prostatectomy.

Given the complexity of the surgical scenes portrayed and the inherent difficulty in detecting surgeon actions, we are confident ESAD will pave the way forward and set a benchmark for the medical computer vision research community. 
The dataset will also help lay the foundations for more robust algorithms to be used in future surgical systems to accomplish various related tasks, such as autonomous assistant surgeons, surgical intervention, surgeon feedback systems, surgical anomaly detection, and so on.


The dataset was released as part of the SARAS-ESAD challenge\footnote{The challenge website is: \url{https://saras-esad.grand-challenge.org}. Please download the dataset from there.} organised at the 2020 Medical Imaging and Deep Learning (MIDL) international conference. The challenge attracted a large number of participants and submissions covering a wide range of approaches. The mean Average Precision (mAP) metric was used to evaluate and rank the methods.
Following standard practice in action detection,
the mAP was computed at three Intersection over Union (IoU) overlap thresholds of 0.1, 0.3 and 0.5. Several submission were able to surpass the performance of the baseline method provided by the organisers. However, only the top ranking methods and a few other approaches which showed significant novelty from an algorithmic point of view are included in this paper (see Section \ref{sec:chal_results}).

\subsection{Paper outline}

The rest of the paper is structured as follows. Section \ref{sec:literature} 
reviews the state of the art, followed in Section \ref{sec:problem} by a problem statement for action detection, in particular in the medical domain. Detailed information on the ESAD dataset, such as the annotation tool and protocols used, the number of action instances per class, etc. are provided in Section \ref{sec:dataset}. The design of the challenge event, the development of the baseline and the evaluation metric are discussed in Section \ref{sec:challenge}. The results of the baseline model follow in Section \ref{sec:results}. The most interesting and best performing methods submitted to the challenge are discussed in Section \ref{sec:chal_results}. The paper concludes with Section \ref{sec:conclusion}.

\section{Literature review} \label{sec:literature}

Action detection or activity analysis from medical images or videos is a field still rather unexplored \cite{sarikaya2020towards}. As a consequence, most work in this field hails in the context of general human gesture and action recognition and/or detection. 
Nevertheless, some relevant efforts deserve to be mentioned.

\subsection{Action and gesture recognition in medical imaging}

Earlier works such as \cite{petlenkov2008application} used the motion of the surgeon's hand to recognise the action performed. Voros~\etal~\cite{voros2008towards} used the motion of the tools to detect the point of interaction between tool and organ. Kocev~\etal~\cite{kocev2014projector} used point clouds generated using a Microsoft Kinect device to build an augmented reality model of the real time actions performed by surgeon, without any attempt to recognise them. 
In~\cite{van2019weakly}, authors used a weakly-supervised approach based on Gaussian Mixture Models (GMM) to recognise surgeon actions. However, the approach was not trained on 
real surgical images (actions were performed on plain uniform artificial surface) and could only recognise actions under the assumption of only one action occurring per frame. 
More recently, Azari~\etal~\cite{azari2019using} used videos of surgeon hand motions to predict surgical maneuvers. The method, however, was limited to actions performed by the surgeon directly with their own hands rather than via teleoperation. 
Li~\etal~\cite{li2016subaction}, developed a temporal action segmentation approach using sub-action categories for early stage prediction. Cascade histogram features were used to identify the starting point of every sub-action. Candidates were then used to train an SVM to classify each sub-action. Again, the algorithm was trained on artificial data, with actions performed manually. 


\subsection{Human action understanding in computer vision}

Within computer vision, the most important distinction is that between \emph{action recognition} (in which the aim is purely to classify a video as an instance of a certain action class), \emph{temporal action segmentation} \cite{derossi2019} (which aims at also deciding when an action starts and stops) and fully-fledged \emph{action detection} (which aims at providing the bounding box(es) around the action(s) of interest, together with the corresponding class scores).

Action recognition methods generally use a 3D convolutional neural network (CNN) to extract spatio-temporal features from the video sequence in order to predict the final action class \cite{ji20123d, tran2015learning, carreira2017quo, singh2019recurrent}. Other approaches use a 2D-CNN to extract spatial features and employ some form of recurrent network to learn the temporal relationships between them \cite{donahue2015long, wang2016temporal, ma2016learning}. 
Temporal action segmentation also requires to extract spatio-temporal information from the video sequence. However, instead of just predicting the class of the action portrayed in the video, these methods also provide the starting and end point of each action instance \cite{singh2016multi, yeung2016end}. 
Finally, action detection methods add even more complexity to the problem by targeting the spatial location of the action being observed \cite{saha2016deep, kalogeiton2017action, peng2016multi}. Such algorithms generally use object detectors to localise the action to later apply some post-processing to link up these spatial action bounding boxes over time to create 'action tubes'.

Another important distinction 
is between \emph{static} and \emph{dynamic} approaches.
Static methods only use spatial information (image data) without any attempt to model the temporal context or dynamics of the action to recognise \cite{singh2017online, gkioxari2015contextual, peng2016multi}. Dynamic activity detection methods, instead, use video data to learn the temporal context of the motion being observed \cite{saha2017amtnet, kalogeiton2017action, hou2017end, hou2017tube}.

\subsection{Action detection}


We would like to mention a few significant examples of work in this area.
Singh~\etal~\cite{singh2017online} used the Single Shot multi-box Detector (SSD) \cite{liu2016ssd} to detect action within each video frame. SSD is capable of predicting an object's bounding box in a single shot, making it one of the fastest object detection algorithms available. Gkiocari~\etal \cite{gkioxari2015contextual} used an R-CNN to output proposal regions potentially containing the action. These proposals were then used to learn the context information to produce a more accurate action class.
Saha~\etal~\cite{saha2017amtnet} proposed an action 
detection module called 3D-RPN (3-Dimensional Region Proposal Network) which uses spatial as well as temporal information from the same video sequence. The model uses two different frames from the same action sequence, separated by an interval of time $\Delta$, to learn the temporal dynamics of the action.
Earlier, Tian~\etal~\cite{tian2013spatiotemporal} had used a deformable part-based model \cite{felzenszwalb2008discriminatively} to detect actions. Peng~\etal~\cite{peng2016multi} developed a motion region proposal network based on the Faster-RCNN architecture \cite{ren2015faster}. There, two streams (RGB images and optical flow) were used to generate action proposals. Jain~\etal~\cite{jain2014action} used supervoxels to generate action bounding boxes. Their method produces $2D+t$ bounding boxes by sampling the videos using Selective Search \cite{uijlings2013selective}.
\\
Kalogeiton~\etal~\cite{kalogeiton2017action} and Hou~\etal~\cite{hou2017tube} both developed action-tube based methods. Both models predict an action tube, which in turn provides spatial bounding boxes in each video frame from the start to the end of an action instance. Li~\etal~\cite{li2018recurrent} proposed a recurrent tubelet proposal and recognition (RTPR) network to address the same task. Their architecture comprises two networks, one for proposal generation and one for recognition, and uses a combination of convolutional and long-short term memory (LSTM) networks to model the recurrent nature of the action proposals.

\subsection{Relevant existing datasets}

\subsubsection{Human action understanding}

While several datasets for action recognition have been released in recent years, action detection research can leverage a comparatively lower number of benchmarks. The largest datasets for action recognition are Kinetics~\cite{kay2017kinetics} and Moments ~\cite{monfortmoments}, the de-facto benchmarks in this area. Notably, the "something-something" dataset \cite{goyal2017something} provides sequences of complex actions performed by humans with everyday objects and comprises 174 fine-grained action classes.
Recently, datasets for temporal activity detection has also been released, including ActivityNet~\cite{caba2015activitynet} and Charades~\cite{sigurdsson2018charadesego}. These datasets are designed to validate methods that detect the start and end point of an action instance within a longer video. 

Relevantly to this work, a few datasets 
have been designed to actually cope with action detection.
Examples are J-HMDB-21 \cite{J-HMDB-Jhuang-2013}, UCF-101-24 \cite{soomro2012ucf101}, LIRIS-HARL \cite{liris-harl-2012} and DALY \cite{daly2016weinzaepfel}, which all provide annotation on the spatial location of each action instance within a video frame together with the start and end time stamp. The more recent AVA dataset \cite{ava2017gu} is currently the largest dataset for action detection with $1.6M$ labeled instances.


\subsubsection{Surgical datasets}

Although no datasets specific to action detection are available in the medical domain, some datasets exist for action recognition, phase recognition, tool detection and other related tasks. In particular, the M2CAI 2016 dataset is divided into a m2cai16-workflow dataset and a m2cai16-tool dataset. The former \cite{twinanda2016endonet} has been developed for the recognition of the phases that compose a cholecystectomy procedure from laparoscopic videos, and contains 41 videos of the procedure. The m2cai16-tool dataset \cite{stauder2016tum} has been developed to validate the detection of tools during cholecystectomy, and contains 15 laparoscopic videos. 
Another relevant dataset is Cholec80, which has been released by the same group for phase recognition and tool detection. This dataset contains 80 videos, again of cholecystectomy procedures \cite{twinanda2016endonet}. 
The Multi-View Operating Room (MVOR) dataset has been recorded using multiple cameras placed at different places in an operating room, and focuses on the 2D and 3D estimation of the pose of operating room staff. The dataset contains 732 synchronised multi-view frames recorded by three RGB-D cameras \cite{srivastav2018mvor}. The JHU-ISI Gesture and Skill Assessment Working Set (JIGSAWS) is a dataset for activity recognition \cite{gao2014jhu} which contains 39 trial samples with synchronised videos and kinematic information targeting a number of standard medical training tasks. Unlike ESAD, however, this dataset was collected in an artificial setup rather than on real MIS procedures.


\section{Problem statement} \label{sec:problem}

\subsection{The action detection problem}

\emph{Action detection} is one of the most complex problems in computer vision. The term is used interchangeably with spatiotemporal action detection. The objective of the task is to identify each action instance in a video sequence by recognising the category of the action being performed, as well as to localise the action instance in both the spatial and the temporal domain. The output is the start and end time of the action instance, 
together with a rectangular bounding box in each video frame between start and stop which identifies where the action is taking place in the image plane.
An action instance can then be represented by a series of bounding boxes linked over time, i.e., an \emph{action tube} \cite{gkioxari2015finding}. 


\subsection{Specificity of action detection from medical images}

As we mentioned, surgeon action detection has peculiarities that make it very challenging. 
The key factor that sets it apart from similar tasks in other domains is the specific nature of the surgical scene and its appearance. 
Among the most important issues are:

\begin{figure}[htb]
    \centering
    \subfloat[\label{fig:p1}]{\includegraphics[width=0.45\textwidth]{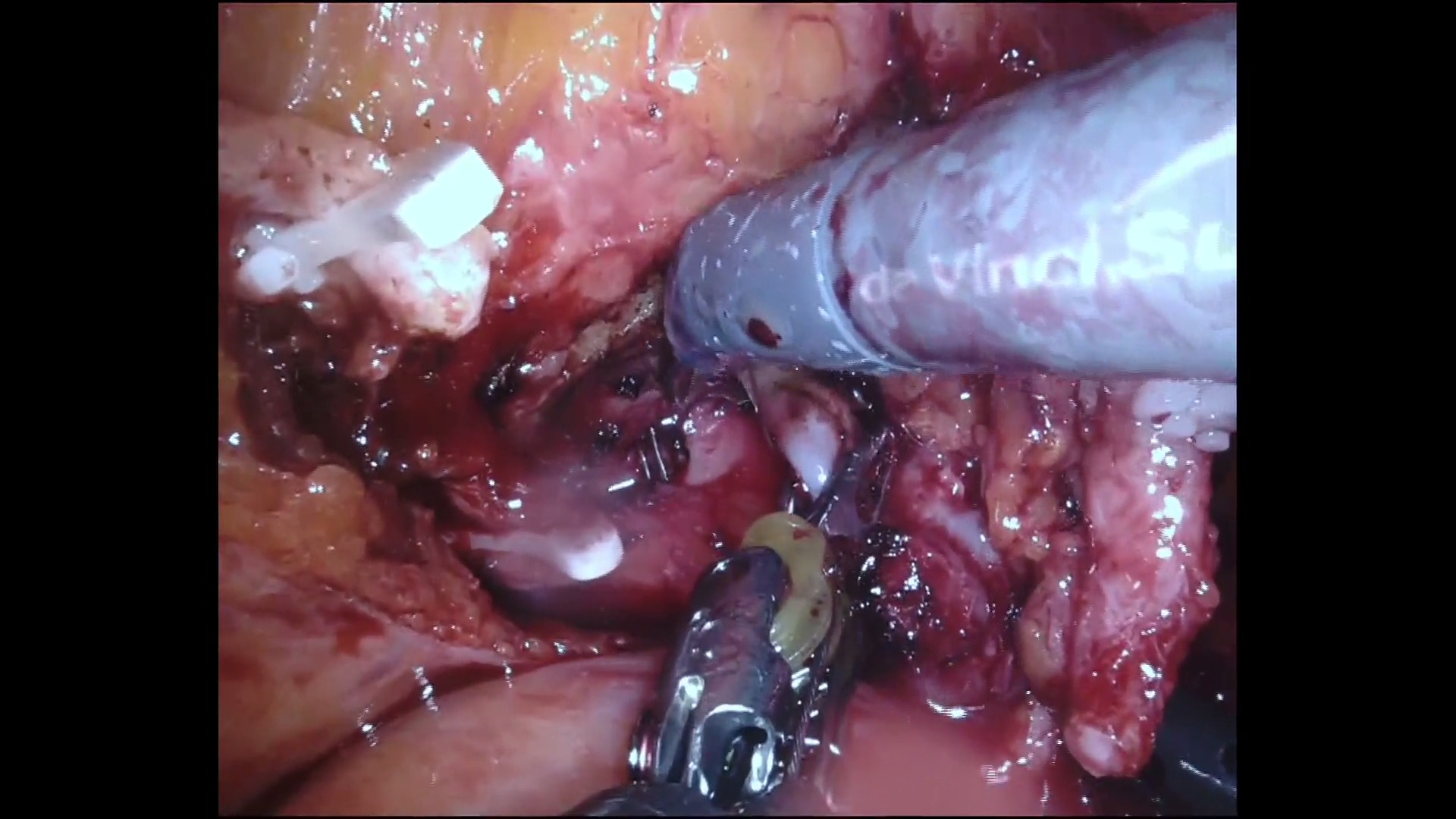}} \hspace{1mm}
    \subfloat[\label{fig:p2}]{\includegraphics[width=0.45\textwidth]{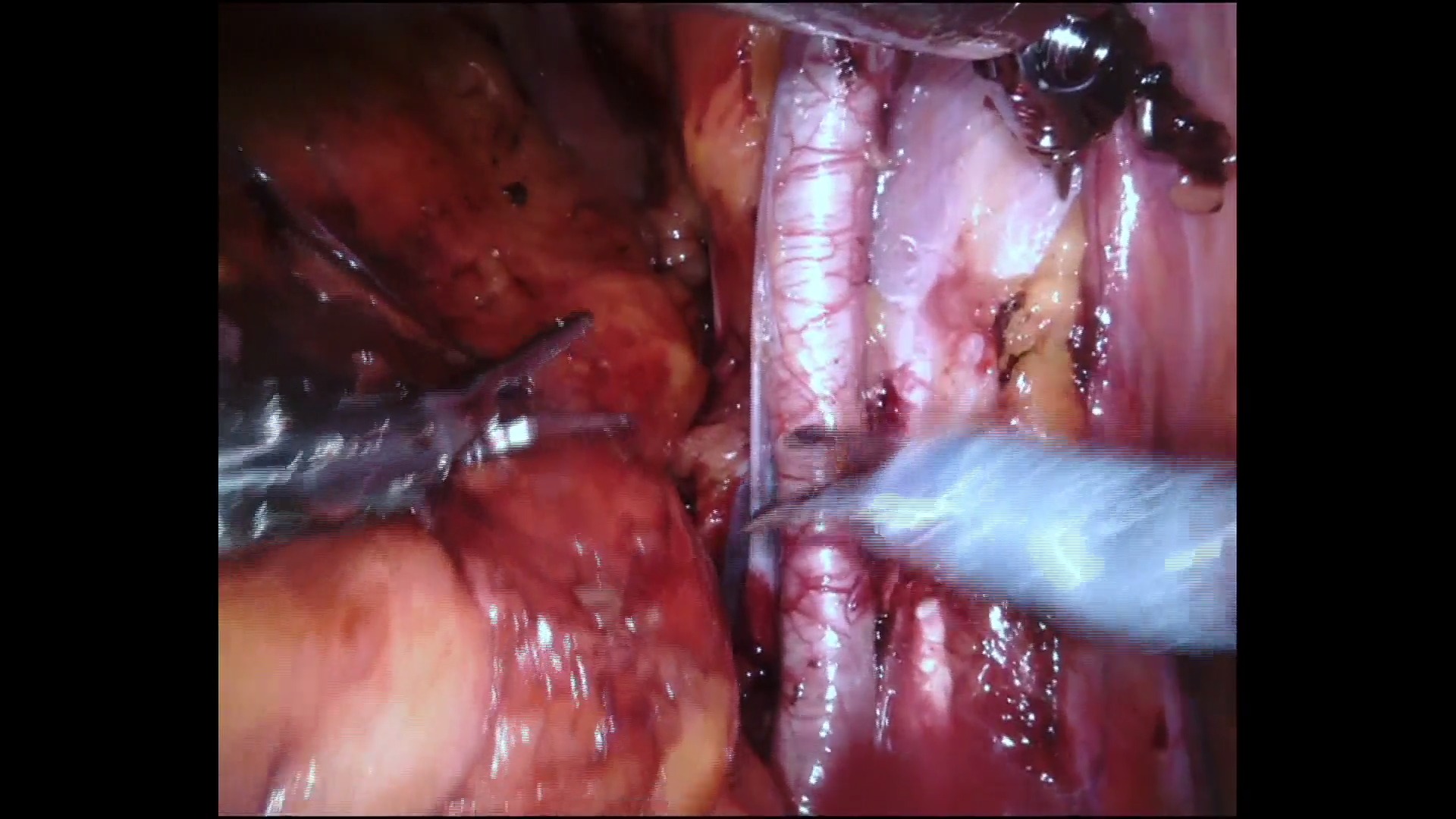}}
    
    \subfloat[\label{fig:p3}]{\includegraphics[width=0.35\textwidth]{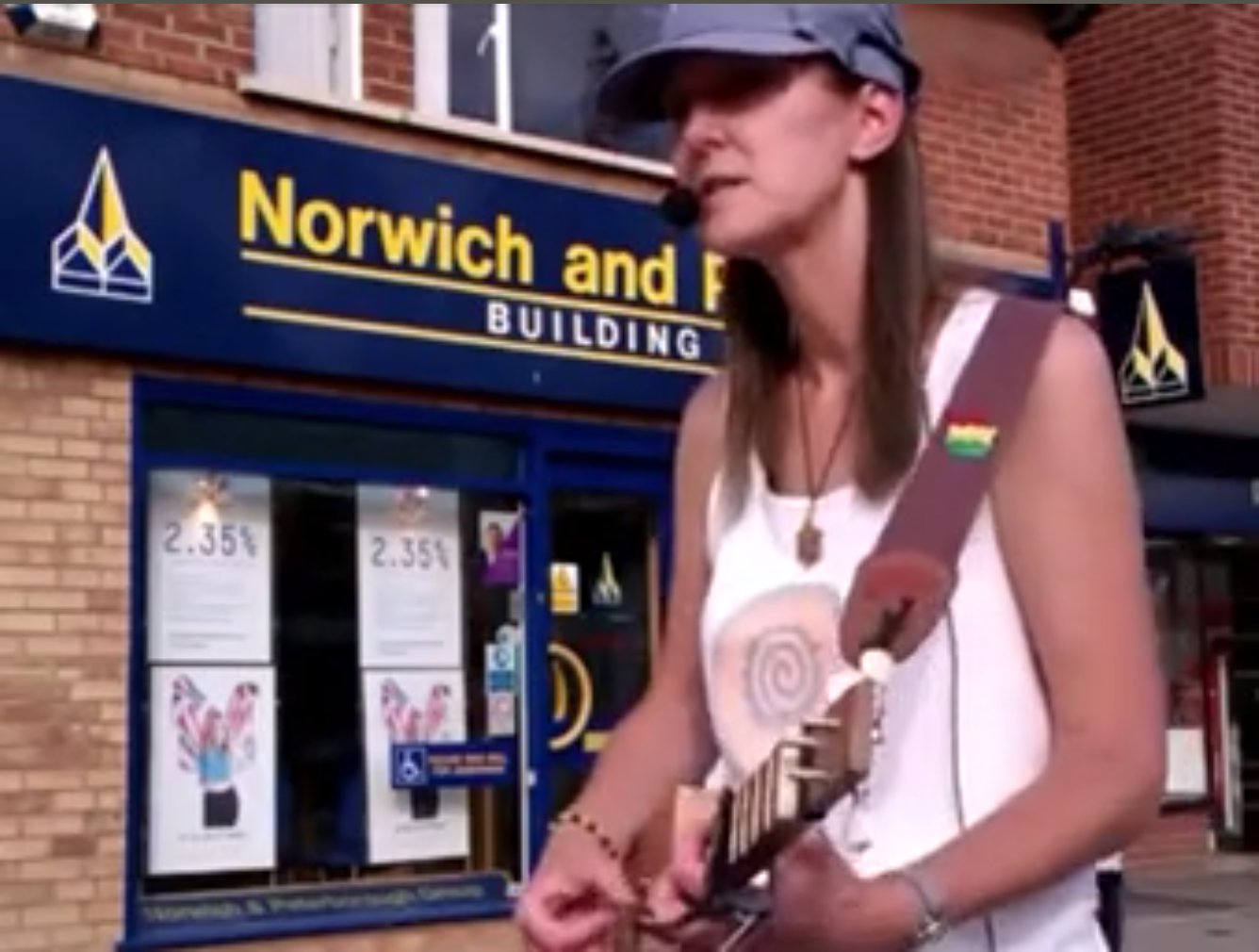}} \hspace{1mm}
    \subfloat[\label{fig:p4}]{\includegraphics[width=0.45\textwidth]{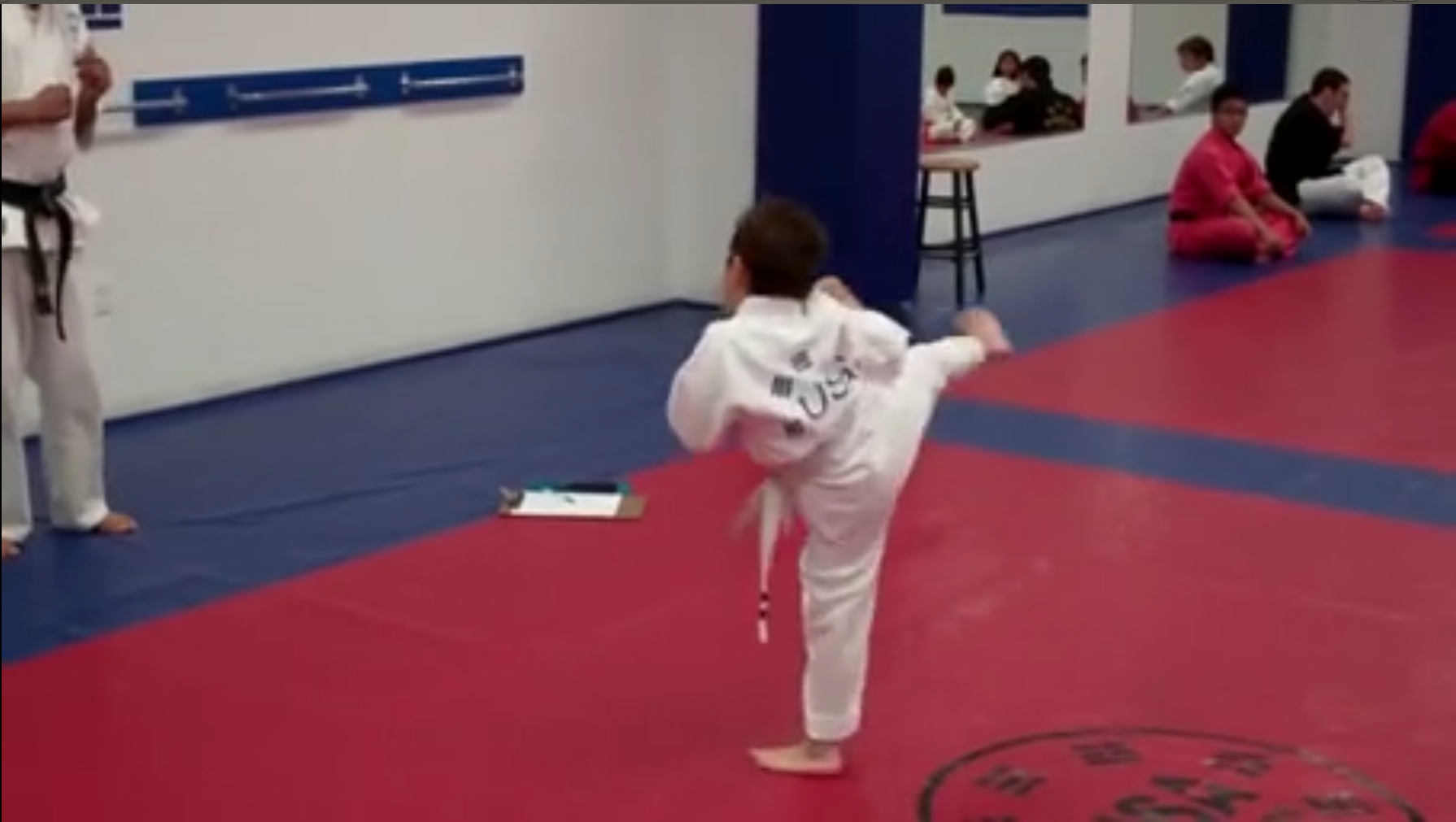}}
    \caption{
    (a) and (b) show sample frames from our ESAD dataset. 
    (c) and (d) portray samples from the Kinetics-400 dataset~\cite{kay2017kinetics}, which collects videos from YouTube. The difference between ESAD and Kinetics~\cite{kay2017kinetics} are evident. In the ESAD dataset, the endoscope captures images from a very close range, thus losing all the useful contextual information human activity videos from YouTube can provide. Additionally, general-purpose action recognition datasets such as~Kinetics~\cite{kay2017kinetics} or AVA~\cite{gu2018ava} provide color, texture, shape and scene information that makes the detection task easier.}
    \label{fig:problem}
\end{figure}

\begin{itemize}
    \item 
    The deformable nature of the organs. As shown in Figure \ref{fig:problem}, organs do not hold a fixed shape in contrast to the traditional human action detection setting, in which the human body has fixed (albeit articulated) shape and is surrounded by objects which also typically have a characteristic shape. 
    \item
    Shape and color variance between two different organs are minimal, and boundaries are hard to recognise, marking a significant difference from standard computer vision tasks (as shown in Figure \ref{fig:problem}).
    \item 
    Endoscopic cameras capture scenes at very close proximity. Hence, the captured frames are unable to show complete organs or their surroundings, delivering little contextual information. In contrast, general activity datasets such as Kinetic~\cite{kay2017kinetics} or AVA~\cite{gu2018ava} allow methods to mine colour, texture, shape and contextual information making it easier to learn discriminative scene features.
    \item 
    The incessant motion and awkward orientation of the endoscope, especially in near proximity, make organs appear very differently from different angles.
    \item 
    The automation of surgical tasks requires to provide a very fine-grained definition of 
    the relevant surgeon actions (e.g., 'CuttingMesocolon', 'PullingProstate' etc.), in which classes differ by small but significant details. 
    It thus becomes highly important to accurately distinguish the organ under operation to be able to accurately detect and predict a surgeon's action.
\end{itemize}

\section{ESAD Dataset} \label{sec:dataset}

This section described first the data collection process (Section \ref{sec:data-collection}), then the annotation protocol we followed (Section \ref{sec:annotation}), to then consider the structure of the resulting ESAD benchmark (Section \ref{sec:structure}). 

\subsection{Data collection} \label{sec:data-collection}

ESAD is composed by four videos of radical prostatectomy procedures conducted on patients, which were collected as part of the EU funded SARAS project by San Raffaele Hospital.
The videos were recorded using a da Vinci Xi robotic system, which comes with an integrated binocular endoscope with a diameter of 8 mm, produced by Intuitive Surgical Inc. Two lenses ($0^o$ or $30^o$) were used. In different stages of the operation, the $30^o$ lens can be used to either look up or down to improve visualisation. The videos used which compose this dataset are monocular.

Sensitive data used by San Raffaele Hospital (images and footage of radical prostatectomies) in the SARAS project was lawfully collected through explicit consent of the data subjects (legal bases: Article 6(1)(a), Article 9(2)(a)) for a previous observational study. The latter was approved, through a specific research protocol, by the Research Ethics Committee of San Raffaele Hospital, composed of more than forty members and currently evaluating more than 30 protocols/month. All surgical videos were anonymised before starting the necessary annotation work.

\subsubsection{Robotic Assisted Radical Prostatectomy}

The four videos depict a Robotic Assisted Radical Prostatectomy (RARP), which is the resection of the whole prostate gland in patients with prostate cancer, with a secondary aim of preserving urinary continence and erectile function. This intervention is the gold standard for robotic-assisted surgery.
The surgical team present in the operating room (OR) was composed by: a main surgeon, operating at the da Vinci console; a surgical assistant (usually a trained urology resident), operating at the surgical table with laparoscopic tools. The duration of RARP surgery is about 3 to 4 hours.
\\
The surgical area was accessed through small incisions in the abdomen and the use of trocars. The first surgeon controlled an advanced robotic system capable of moving surgical tools from outside the body. A high-tech interface would let the surgeon use natural wrist movements and a 3D screen during the entire operation. One of these trocars was placed over the umbilicus for camera port. This involves inserting a fibre-optical instrument and some other operating instruments into the patient’s inflated abdomen. The camera would stream video data to the operator’s console, where it was used to have a view of the patient’s abdomen.

Two possible approaches exist as to how to access the surgical area within RARP: (i) the transperitoneal approach, with access to the abdomen, and the (ii) extraperitoneal one, with pelvic access. Most RARPs are executed through the transperitoneal approach, which is indeed the situation described in the SARAS procedural workflow.
Within the transperitoneal approach itself, two different modalities for reaching the target organs during RARP exist. In the anterior modality, after transperitoneal access and insufflation, the space of Retzius is immediately entered and the prostate gland, seminal vesicle, and vasa are reached and dissected from the front. In the posterior modality, the seminal vesicles and vasa are initially reached and completely dissected behind the bladder.
The videos selected for ESAD concerned the RARP posterior approach, for this procedure is routinely performed in the clinical practice by the expert urological surgeons of San Raffaele Hospital.

Note, however, for the purpose of simplifying the RARP procedure to be implemented in the SARAS project demonstrator, it was necessary to select the transperitoneal anterior approach. This change was dictated by pre-testing evidence on the robotic platform and phantoms and, in accordance with San Raffaeke surgeons, was aimed to enlarge the surgical working space and to optimise the anatomical reconstruction of the phantom. 

\subsection{Annotation protocol} \label{sec:annotation}

For action detection purposes, the videos need to be annotated by manually providing bounding boxes around the actions of interest in each video frame, and by inputting the class label associated with each bounding box.

\subsubsection{The Virtual object Tagging Tool}

\begin{figure}[htb]
    \centering
    \includegraphics[width=\textwidth]{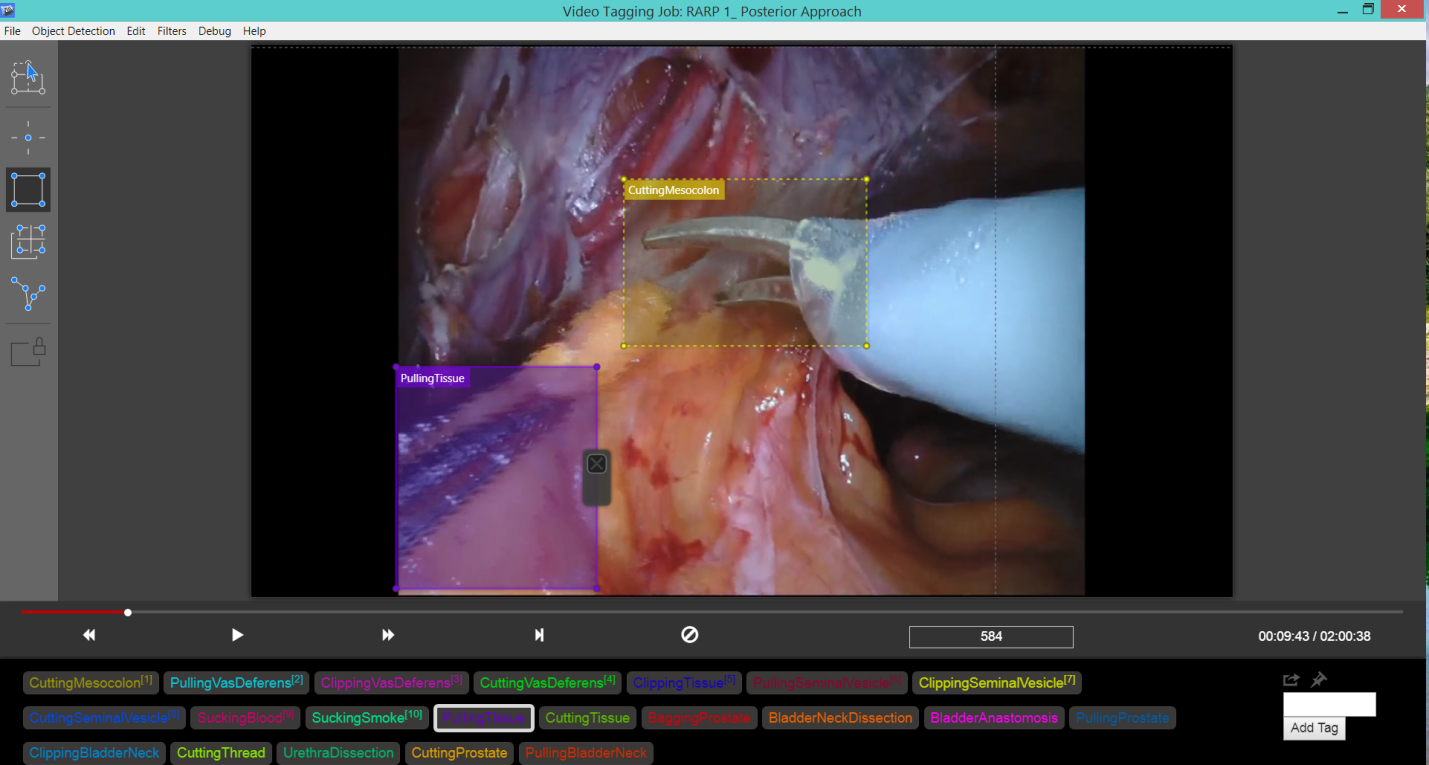}
    \caption{Screenshot captured while using VoTT for ESAD annotation.}
    \label{fig:vott}
\end{figure}

Video annotation was performed using the Virtual object Tagging Tool (VoTT). VoTT is a Microsoft open source tool used for drawing bounding boxes around regions of interest in visual data. 
A screenshot of its graphical user interface is shown in Figure \ref{fig:vott}

\subsubsection{Issues with the annotation process}

The annotation process raises a number of issues. To begin with, what constitutes an action or an event of interest is somewhat unclear. Some researchers have considered for this purpose surgical tools tracking methodologies [23], but for action detection and classification this would evidently cause the model to focus too much on the tools, rather than on what happens in the surgical cavity. As a result, the model would detect many false actions whenever a tool appears in the field of view. The same could happen when focusing on tissue strands or organs. We therefore decided to explore a combination of both organs and tools when setting the list of actions of interest and their descriptions. As a result, bounding boxes were drawn only when tools were close to the appropriate organs in order to deliver the identified actions of interest.
\\
What is the ideal size of a bounding box is also unclear. To balance the presence of tools and organs or tissue in a bounding box, bounding boxes were restricted to containing 30\%-70\% of either tools or organs. 
\\
Last but not least, the annotation protocol needs to precisely specify rules for determining the temporal extent of each action.

Overall, the annotation task is subject to the inherent ambiguity of discriminating visually similar classes. For instance, it is hard to tell whether the aspirator is sucking blood, pushing some organs to make way, or sucking smoke. This was mitigated by seeking expert knowledge.

\subsubsection{Annotation guidelines}

A set of protocols was developed to guide the annotators in their work. This helped minimise the ambiguity in deciding the size of bounding boxes around each action instance, as well as their locations. All annotators were provided a set of instructions with examples in order to standardise the procedure as much as possible.
The following guidelines were enforced:

\begin{itemize}
    \item Each bounding box should contain both the organ and tool performing the action under consideration, as each action class is highly dependent on the organ under operation.
    \item To balance the presence of tools and organs or tissue in a bounding box, bounding boxes are restricted to containing 30\%-70\% of either tools or organs.
    \item An action label is only assigned when a tool is close enough to the appropriate organ, as informed by the medical expert. Similarly, an action stops as soon as the tool starts to move away from the organ.
\end{itemize}

\subsection{Characteristics of the dataset} \label{sec:structure}

\subsubsection{List of actions}

After a rigorous analysis of the actions actually performed by the surgeons in the four available RARP videos, we selected 21 action categories for inclusion in ESAD. The decision was made while keeping in mind that action categories should not be so simple that they cannot provide any useful information (e.g., for an autonomous decision making module as in the SARAS architecture). Similar problems were encountered in previous medical action recognition datasets \cite{azari2019using, petlenkov2008application}. At the same time the action classes should not be so complex as to make it impossible to perform the task. 

The final list of action classes was decided with the help of multiple surgeons and medical professionals. 
The list is shown in Table \ref{tab:dataset}, along with the number of action instances for each category in the whole dataset. 

\begin{table}[!htb]
    \centering
    \caption{List of actions in the ESAD dataset, with the number of samples in each of the training, validation and test folds.}
    \label{tab:dataset}
    \begin{tabular}{lcccc}
    \toprule
          Label & Train & Val & Test & Total instances \\
    \midrule 
         CuttingMesocolon & 315 & 179 & 188 & 682 \\
         PullingVasDeferens & 457 & 245 & 113 & 815 \\
         ClippingVasDeferens & 33 & 25 & 48 & 106 \\
         CuttingVasDeferens & 71 & 22 & 36 & 129 \\
         ClippingTissue & 215 & 44 & 15 & 274 \\
         PullingSeminalVesicle & 2712 & 342 & 436 & 3490 \\
         ClippingSeminalVesicle & 118 & 35 & 33 & 186 \\
         CuttingSeminalVesicle & 2509 & 196 & 307 & 3012 \\
         SuckingBlood & 3753 & 575 & 1696 & 6024 \\
         SuckingSmoke & 381 & 238 & 771 & 1390 \\
         PullingTissue & 4877 & 2177 & 2024 & 9078 \\
         CuttingTissue & 3715 & 1777 & 2055 & 7547 \\
         BaggingProstate & 34 & 5 & 37 & 76 \\
         BladderNeckDissection & 1621 & 283 & 519 & 2423 \\
         BladderAnastomosis & 3585 & 298 & 1828 & 5711 \\
         PullingProstate & 958 & 12 & 451 & 1421 \\
         ClippingBladderNeck & 151 & 24 & 18 & 193 \\
         CuttingThread & 108 & 22 & 40 & 170 \\
         UrethraDissection & 351 & 56 & 439 & 846 \\
         CuttingProstate & 1845 & 56 & 48 & 1949 \\
         PullingBladderNeck & 189 & 509 & 105 & 803 \\
    \bottomrule
    \end{tabular}
\end{table}

\subsubsection{Duration, sampling rate}

On an average, each of the four RARP videos which comprise ESAD is 2 hours and 20 minutes long. Each video is recorded at 30 frames per second (FPS), whereas the annotation is performed at 1 FPS to avoid long stretches in which the scene changes very little. 
Each frame can contain more than one action instances. Each instance is annotated using a bounding box and its action label from the list of classes. 
As during the procedure tools operate in close proximity, the dataset contains many instances of action with overlapping bounding boxes. 

\begin{figure}[htb]
\center
\subfloat{\includegraphics[width=.49\textwidth]{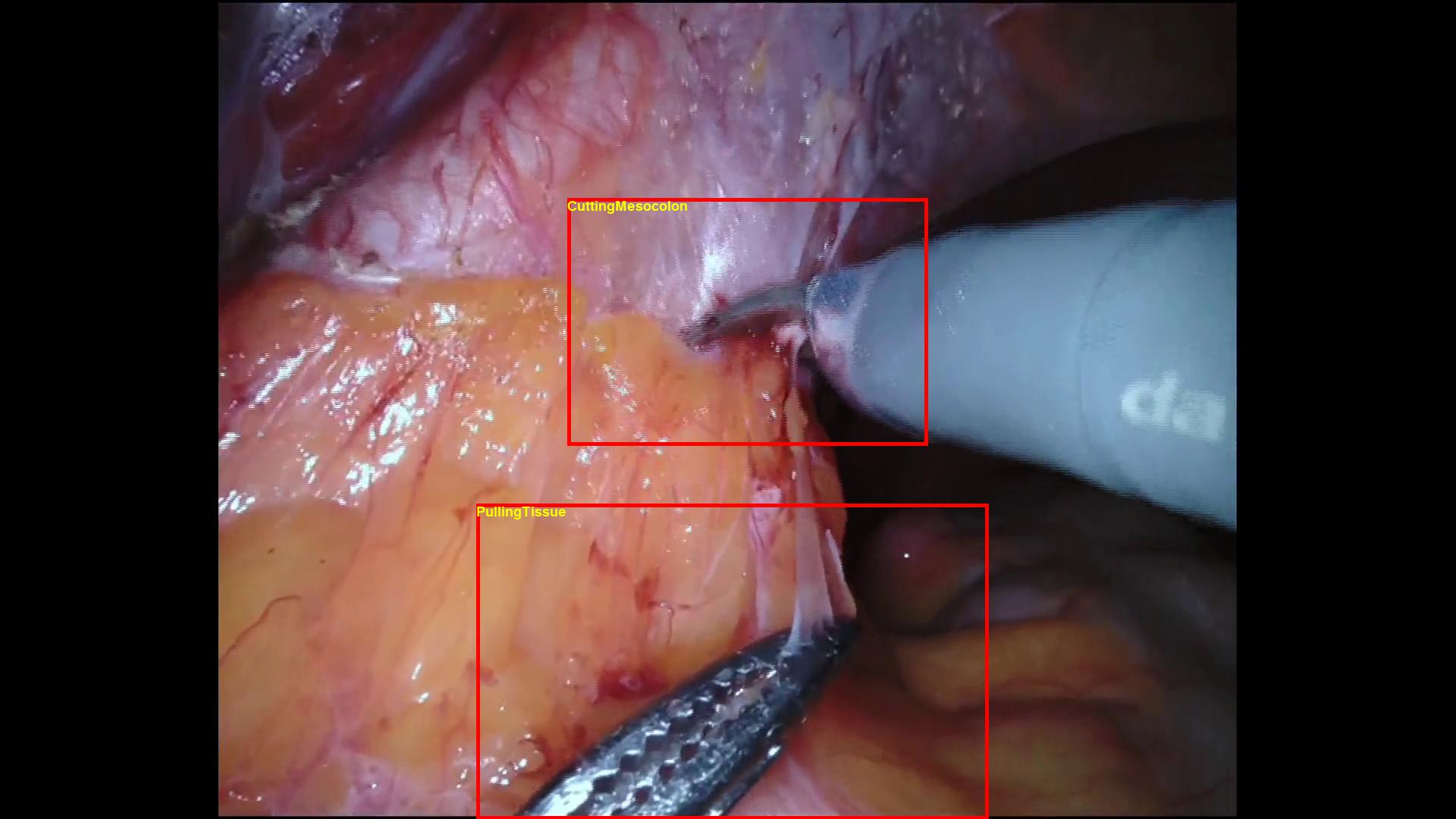}} \hspace{1mm}
\subfloat{\includegraphics[width=.49\textwidth]{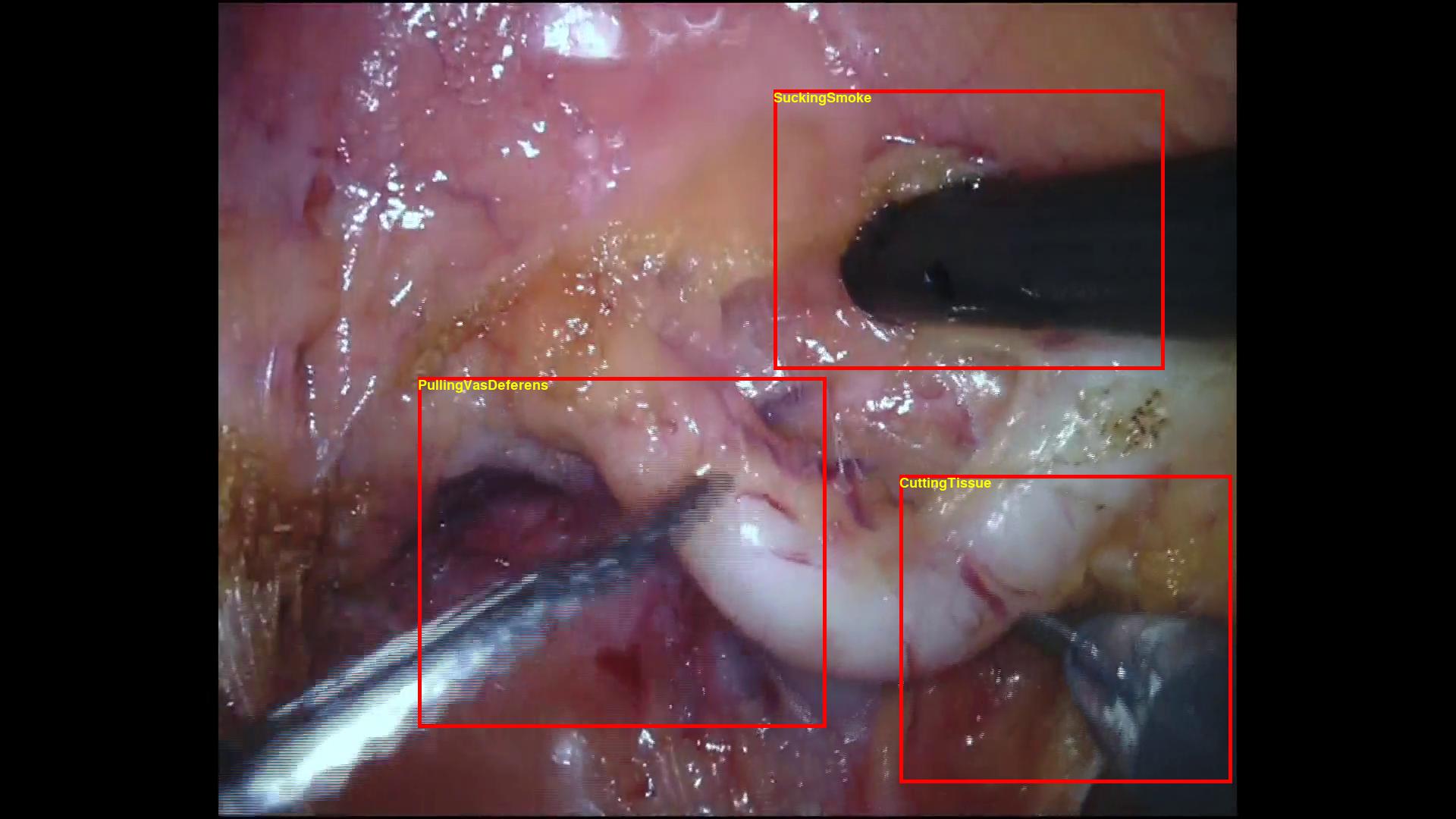}}

\subfloat{\includegraphics[width=.49\textwidth]{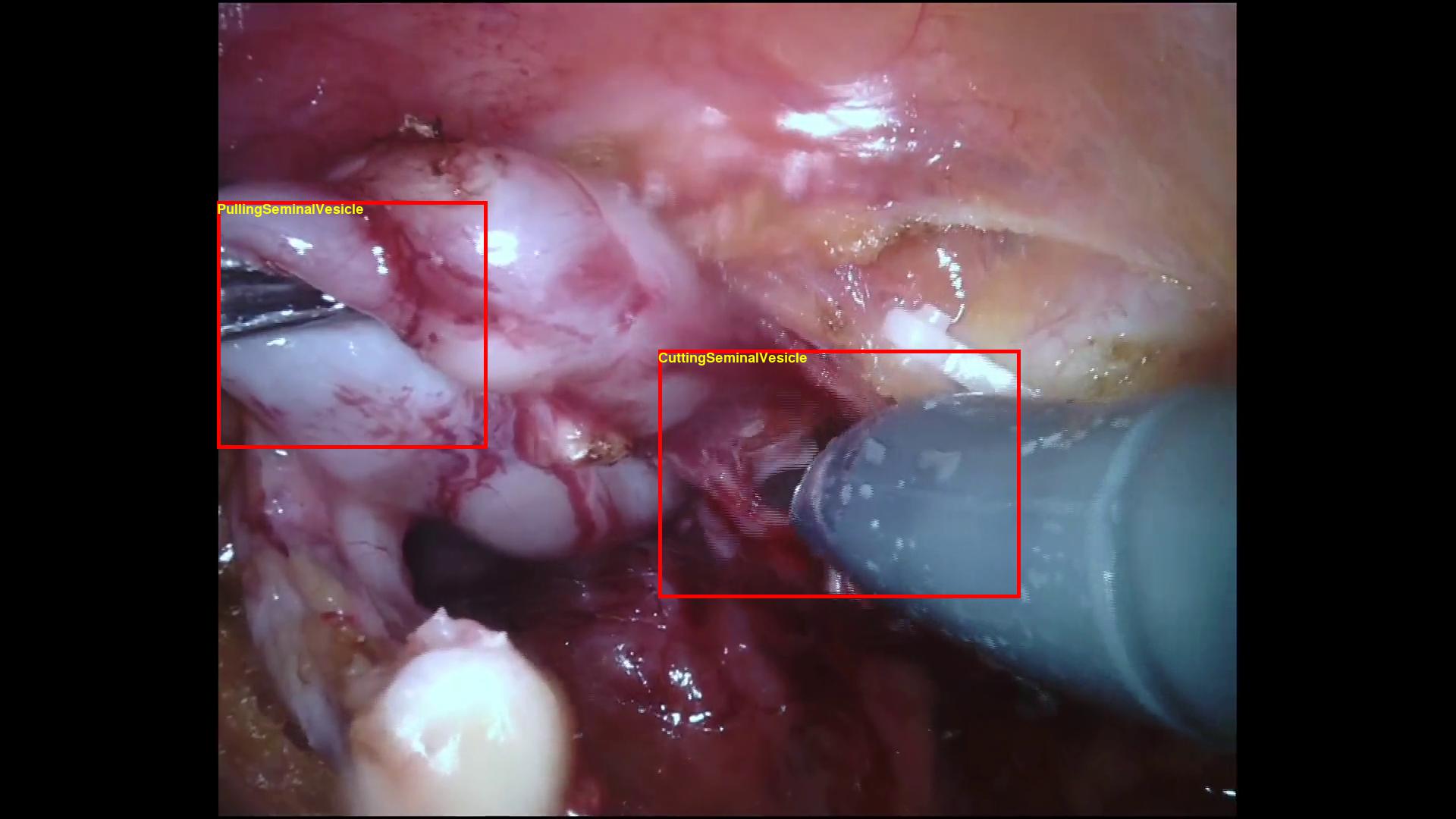}} \hspace{1mm}
\subfloat{\includegraphics[width=.49\textwidth]{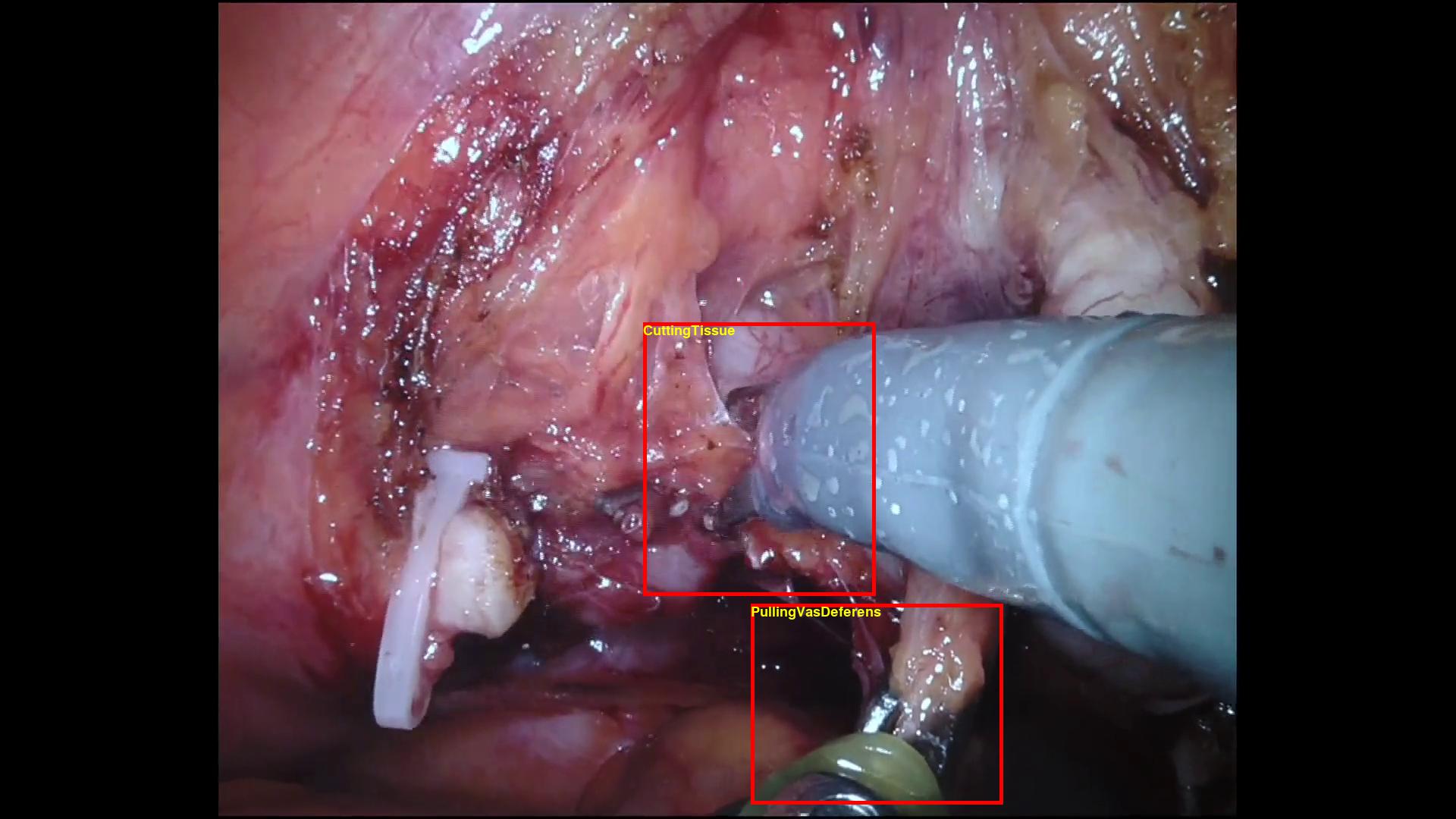}}
\caption{Sample video frames from the ESAD dataset, with the corresponding surgeon actions annotations. A red box in figure denotes a bounding box. The video frames are captured by the endoscope during a prostatectomy procedure.}
\label{fig:anno_image}
\end{figure}

Some example video frames with annotation are shown in Figure \ref{fig:anno_image}. From the images it can been appreciated that all bounding boxes are centred around a tool, as laparoscopic tools represent the 'subject' of the action, but we make sure to include a portion of the organ undergoing the operation. 

The reason for that is that most surgical actions have different names depending on the organ being operated upon, despite having in common the same motion pattern of the same tool. This results in a very fine-grained recognition problem, as it can be appreciated from Table \ref{tab:dataset}.

\begin{figure}[htb]
    \centering
    \includegraphics[width=\textwidth]{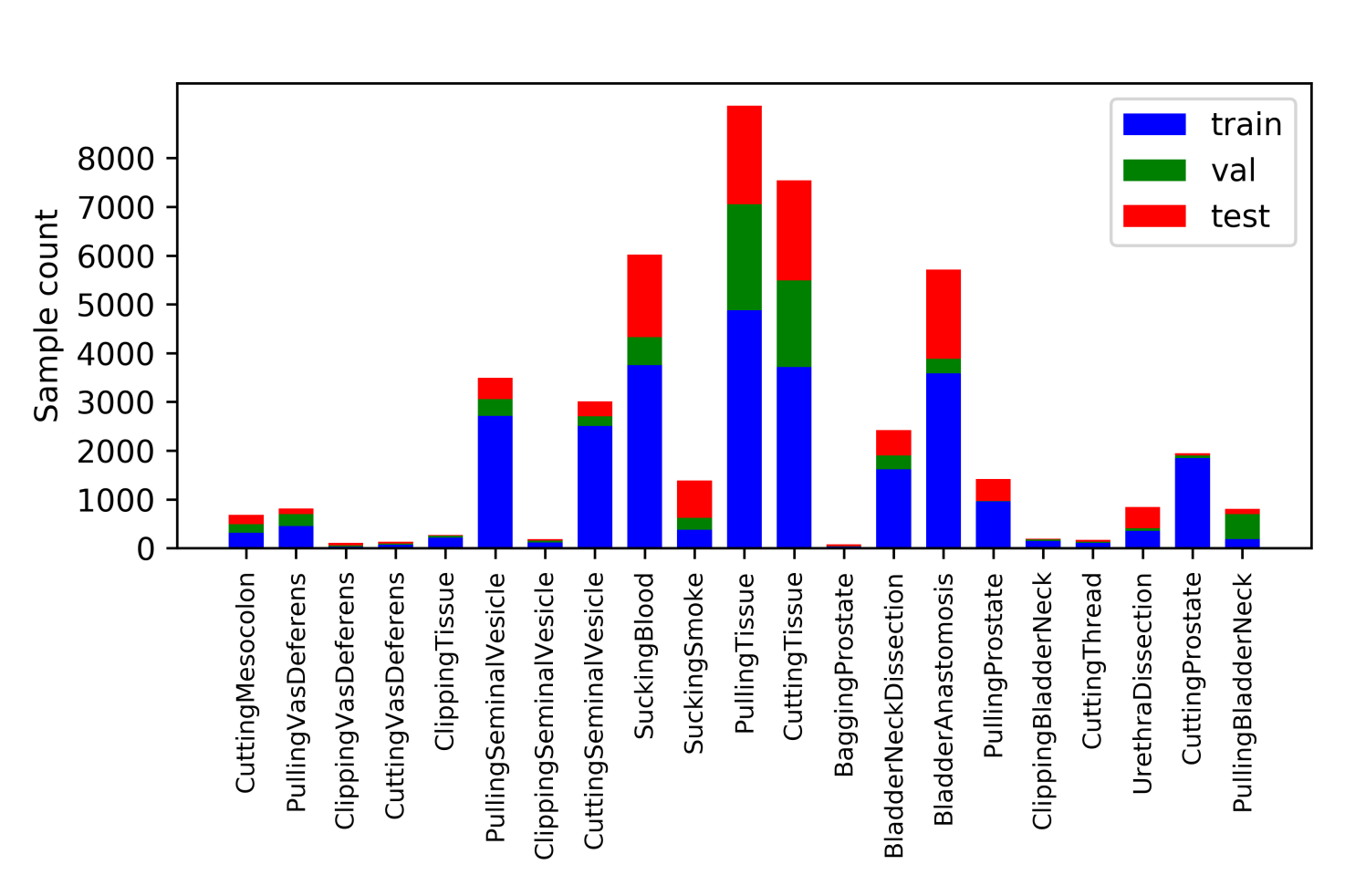}
    \caption{Distribution of the number of samples per action category in the training, validation and test sets, represented by blue, green and red bars in the diagram, respectively. \label{fig:esad_dist}}
\end{figure}



\section{Challenge} \label{sec:challenge}


\subsection{Training and evaluation split} \label{sec:split}

We briefly recall the protocol for splitting the ESAD dataset into training and evaluation folds we adopted for the MIDL 2020 SARAS-ESAD challenge.

The dataset was divided into three different sets: training, validation and test. The two procedures with the highest number of action instances were selected to form the training set. The one video with the most balanced number of samples for each action class was used as the test set. 
The goal was to provide a fair evaluation of all possible types of algorithms, under a limited but realistic modelling of the domain variations in the problem. The last procedure was selected as the validation set. 
The training set contained a total of 22,601 annotated frames, comprising 28,055 action instances. 
The validation set contained 4,574 frames portraying 7,133 action instances in total. The test set (released only in a second moment), contained 6,223 annotated frames with 11,565 action instances.

The distribution of the samples for each action category and for each of the three splits is shown in Figure \ref{fig:esad_dist}. It is clear from the chart that ESAD is highly skewed in term of class imbalance. The reason for this is the nature of surgical procedures itself, and of RARP in particular. As shown in Figure \ref{fig:esad_dist}, classes \textit{PullingTissue} and \textit{CuttingTissue} contain the largest number of samples, as these are the most common actions performed by a surgeon during prostatectomy. In opposition, classes such as \textit{BaggingProstate} and \textit{CuttingThread} have the lowest number of samples, because of the short duration of these activities in the course of each procedure. 

\subsection{Challenge task}

For our MIDL 2020 SARAS-ESAD challenge we decided to only set one task, namely that of action detection from individual images (video frames), rather than ask the participants to also tackle the more difficult challenge of video-based action detection (i.e., the detection of series of bounding boxes delimiting an action instance or ’action tube’). The main reason for this decision was that this is a very new challenge for medical computer vision community and the task is already complex enough.
Challenge entries were thus required to predict the class as well as the bounding box (location) of the action instances present in any given frame. The presence of multiple simultaneous actions in a single frame further complicated the problem.

We do plan to augment the challenge in the near future by adding more tasks and asking for video detection evaluation.

\subsection{Evaluation metrics}
\label{subsec:eval-metric}

\subsubsection{Average precision}

Detection tasks are normally assessed in terms of \emph{average precision} (AP), which does not merely focus on the percentage of misclassified examples. For each class, AP requires to calculate both the percentage of instances correctly classified as positive over the total of those classified as positive (\emph{precision}) and the rate of positive instances correctly recognised as such, known as \emph{recall}. 
Namely, we define \emph{precision} and \emph{recall} as follows:
\[
Precision = \frac{TP}{TP + FP}, \quad Recall = \frac{TP}{TP + FN},
\]
where TP is the number of true positives, FP is the number of false positives and FN is the number of false negatives.
Average precision is then obtained by plotting precision against recall (yielding a precision-recall curve) and then integrating the area under the curve.

\emph{Mean Average Precision} (mAP) over a set of query points is the mean of the AP scores for each query. For our action detection problem, one can define the \emph{Frame-mAP} value as the mean of the AP values for each of the individual frames. 

\subsubsection{IoU thresholds}

Whether a predicted detection is correct or not can be measured by the percentage overlap between the predicted and ground truth bounding boxes. The \emph{Intersection over Union} (IoU) value is the ratio between the area of the intersection of the two bounding boxes over the area of their union.
In our challenge we used three different IoU thresholds to assess the correctness of the detections and to compute the corresponding average precision.
We computed the average precision at IoU = 0.1, 0.3 and 0.5, respectively, obtaining figures we call in the following $AP_{10}$, $AP_{30}$ and $AP_{50}$. Their mean was also computed to get a final evaluation score for the challenge. 

The purpose of computing AP at three different degrees of detection overlap is to capture the quality of the detection as well as that of the classification.
As we know, this is a new and challenging task, for which it is very difficult to get good detection performance at higher thresholds (as can be observed in the $AP_{50}$ column of our baseline results, see Table \ref{tab:esad_results_ResNet50}). Hence, we wanted to assess both how accurately a model can detect the classes of action present in the scene as well as how precise is the location of the predicted bounding boxes. 

\subsection{Baseline model}~\label{subsec:baseline_model}

We released a baseline model before the start of the challenge to provide a reference for the competing teams. 

\subsubsection{Backbone}

Our baseline model is based on a Feature Pyramidal Network (FPN) architecture, a concept originally proposed by Lin \etal~\cite{lin2017feature} which uses a CNN architecture with pooling layers. 
A residual network (ResNet)~\cite{he2016deep} is used as the `backbone' network for the detection model: the output of each residual block is used to build the feature pyramid. The residual feature maps at different levels of pyramid are then fed to a sub-net composed of four convolutional layers. Finally, an additional convolutional layer predicts the class scores and bounding box coordinates, respectively. 

Similarly to what done in the original paper~\cite{lin2017feature} we freeze the batch normalisation layers of the ResNet-based backbone. In addition, a few initial layers are also frozen to avoid overfitting. Finally, non-maximal suppression (NMS) \cite{rothe2014non} 
is used to discard the false positives among model predictions at test time.

\subsubsection{Losses}

We tried two different loss functions to train the classification sub-net of our baseline model. Since the latter is a single-stage model based on \cite{lin2017feature}, following the original paper we trained our FPN with both an online hard example mining (OHEM)-loss \cite{liu2016ssd} and focal loss \cite{lin2017focal}. The OHEM loss 
assigns a larger weight to more difficult (`hard') examples than to easier ones. Focal loss, on its part, is specifically designed to handle class imbalance in the training data. As class imbalance is a critical feature of this dataset, we believe it is important to see how focal loss affects the performance of the detector. 
To train the regressed bounding boxes coordinates in the sub-net we
used a smooth $l_1$ loss~\cite{ren2015faster}.

\subsubsection{Implementation details} 

We trained our baseline model using various input image sizes, namely 200, 400, 600 or 800 pixels (short, row side), by rescaling the original video frame while preserving its aspect ratio. 
For training, we set the learning rate to $0.01$ and the batch size to $16$. The networks were trained for $7K$ iterations with a learning rate drop of a factor of $10$ after $5K$ iterations.

The complete model was implemented in \href{https://pytorch.org/}{pytorch} and is provided open access\footnote{The baseline model is available at: \url{https://github.com/Viveksbawa/SARAS-ESAD-Baseline}}. 
At the moment, the source code supports PyTorch 1.5 and Ubuntu with the Anaconda distribution of python. It was tested on machines with 2, 4 and 8 GPUs.

\section{Baseline results}~\label{sec:results}


The results achieved by our baseline model on the ESAD dataset, using both loss functions, are shown in Table \ref{tab:esad_results_ResNet50}. 
As mentioned we trained the model with four different image sizes in order to observe the effect of the size of the surgical tools on detection accuracy.
As we can observe in Table \ref{tab:esad_results_ResNet50}, as the image size increases models with an OHEM loss function are able to achieve better detection accuracy on the validation set. The same, however, cannot be said for the test-set. We believe that this performance gap 
is due to the lower sample count for some classes in the training data. Moreover, the test set has lower class imbalance than the validation set. This also highlights the issue of scene complexity in medical images, as the latter have very high intra-class variation and very little inter-class variation. This issue is further (involuntarily) exhacerbated by our choice of action classes, whose meaning is highly dependent on the organ being manipulated.
\\
Table \ref{tab:esad_results_ims400} shows the results achieved by our baseline model using different backbone networks, while fixing the input image (row) size to 400, on both validation and test sets. 


\begin{table}[!htb]
    \centering
    \caption{Results of the baseline models with different loss functions and input image sizes. The backbone network is set to ResNet50. The columns $AP_{10}$, $AP_{30}$, $AP_{50}$ report the average precision at IoU threshold values of 0.10, 0.30 and 0.50, respectively, on the validation set. The column $AP_{mean}$ and $Test-AP_{mean}$ report the mean average precision computed by taking the mean of three AP values on the validation and test sets, respectively. 
    }
    \label{tab:esad_results_ResNet50}
    \begin{tabular}{lcccccc}
    \toprule
        Loss & Row size & $AP_{10}$ & $AP_{30}$ & $AP_{50}$ & $AP_{mean}$ & $AP_{mean} (Test)$ \\ \midrule
        Focal & 200 & 33.8 & 17.7 & 6.6 & 19.4 & 15.7 \\
        Focal & 400 & 35.9 & 19.4 & 8.0 & 21.1 & \textbf{16.1} \\
        Focal & 600 & 29.2 & 17.6 & 8.7 & 18.5 & 14.0 \\
        Focal & 800 & 31.9 & 20.1 & 8.7 & 20.2 & 12.4 \\
        OHEM & 200 & 35.1 & 18.7 & 6.3 & 20.0 & 11.3 \\
        OHEM & 400 & 33.9 & 19.2 & 7.4 & 20.2 & 13.6 \\
        OHEM & 600 & 37.6 & 23.4 & 11.2 & 24.1 & 12.5 \\
        OHEM & 800 & \textbf{36.8} & \textbf{24.3} & \textbf{12.2} & \textbf{24.4} & 12.3 \\
        \bottomrule
    \end{tabular}
\end{table}

\begin{table}[!htb]
    \centering
    \caption{Results of the baseline models with different loss functions and backbone networks, with input image size set to 400. Columns $AP_{10}$, $AP_{30}$, $AP_{50}$ report the average precision at IoU threshold values of 0.10, 0.30 and 0.50, respectively, on the validation set. Columns $AP_{mean}$ and $Test-AP_{mean}$ report the mean average precision computed by taking the mean of the three AP values on the validation and test sets, respectively.}
    \label{tab:esad_results_ims400}
    \begin{tabular}{lcccccc}
    \toprule
    Loss & Backbone & $AP_{10}$ & $AP_{30}$ & $AP_{50}$ & $AP_{mean}$ & $AP_{mean} (Test)$ \\ \midrule
    Focal & ResNet18 & 35.1 & 18.9 & 8.1 & 20.7 & \textbf{15.3} \\
    OHEM & ResNet18 & 36.0 & \textbf{20.7} & 7.7 & \textbf{21.5} & 13.8 \\
    Focal & ResNet34 & 34.6 & 18.9 & 6.4 & 19.9 & 14.3 \\
    OHEM & ResNet34 & \textbf{36.7} & 20.4 & 7.1 & 21.4 & 13.8 \\
    Focal & ResNet50 & 35.9 & 19.4 & \textbf{8.0} & 21.1 & 16.1 \\
    OHEM & ResNet50 & 33.9 & 19.2 & 7.4 & 20.2 & 13.6 \\
    Focal & ResNet101 & 32.5 & 17.2 & 6.1 & 18.6 & 14.0 \\
    OHEM & ResNet101 & 36.6 & 20.1 & 7.4 & 21.3 & 12.3 \\
    \bottomrule
    \end{tabular}
\end{table}

It is clear from Tables~\ref{tab:esad_results_ims400} and~\ref{tab:esad_results_ResNet50} that the OHEM loss performs better on the validation set, while focal loss performs better on the test set. While theoretically focal loss is designed to handle the problem of class imbalance, empirically it did not prove to be very effective. There is no clear pattern in the performance of the model in terms of the choice of a loss function. On the basis of the formulation of the two losses, we can safely claim that the simpler weighting mechanism of OHEM helps maintaining sufficiently large gradient magnitudes throughout the learning process, leading to better convergence. 

If we look at the dependency of performance from the backbone, lower-depth networks seem to provide better detection results. ResNet18-based models achieve the highest mAP on both the validation set and the test set. Again, this might be due to the fact that a lower depth and a lower parameter count reduce the probability of overfitting to the training data, leading to better generalisation power. Overall, it is clear that the presented baseline method is still far from achieving a satisfactory performance, certainly not allowing the real-world deployment of such as system, with a mean average precision generally hovering around 20\%.

\section{Results of the challenge} \label{sec:chal_results}

\begin{figure}[htb]
    \centering
    \includegraphics[width=\textwidth]{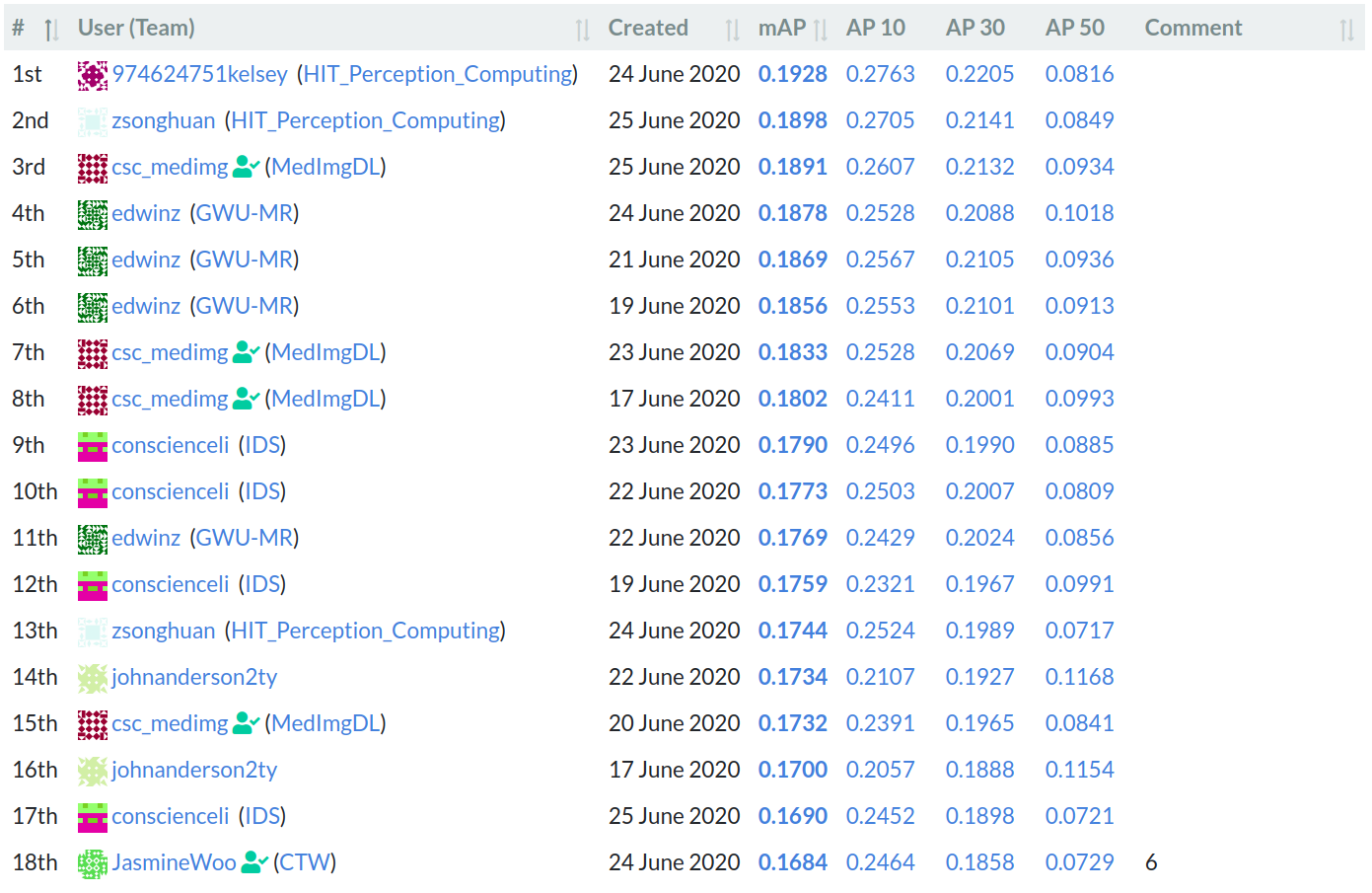}
    \caption{Snapshot of the leaderboard from the challenge website.}
    \label{fig:my_label}
\end{figure}

\subsection{Overview: single-stage versus two-stage detectors}

A large number of entries were submitted to the challenge ($75$, from more than $20$ different research teams: see Fig. \ref{fig:my_label} for the top of the leaderboard). The methods can be generally divided into \emph{single-stage} versus \emph{multi-stage} action detection approaches.

Single-stage algorithms jointly predict the location of the action as well as its category at the same time. Popular examples of single-stage detectors are the Single Shot Detector (SSD) \cite{liu2016ssd} and You Only Look Once (YOLO) \cite{redmon2016you}, among others. 
Two-stage approaches, instead, use two sub-modules. While the first stage is used to localise the action, the second stage predicts the action class. 
Examples of two-stage object detectors are, among others, R-CNN \cite{girshick2015fast} and Faster R-CNN \cite{ren2015faster}. 
In general, single-shot detectors are considered computationally more efficient and faster at inference time when compared to two-stage models, but the latter are typically able to provide better detection accuracy. 
Although this gap in performance 
has decreased over time for mainstream task and datasets \cite{zhang2020bridging, li2018exploring, zhao2019object}, such as Microsoft COCO \cite{lin2014microsoft}, 
for more complex tasks (such as action detection) two-stage detectors still appear to outperform single-stage models.
\\
This hypothesis was also validated by the results of our challenge. In the following we structure the discussion by topic, rather than by entry.

\subsection{Changing the backbone and attention mechanisms}

Among the top performers, some submissions, which we describe here first, chose to modify the baseline model by using different backbone and attention mechanisms. 
Table \ref{tab:results_1} presents the impact of backbone replacement when only the training data are used to train the model, while evaluation is performed on the test set. The table presents the change in model performance associated with seven different backbones, based on ResNeSt \cite{zhang2020resnest}, CBAM \cite{woo2018cbam} and a traditional ResNet architecture. All backbones used models pretrained on the Imagenet dataset, as is standard practice in computer vision. For all these models focal loss with input image size of $200 \times 360$ were used for both training and evaluation. 

Both ResNeSt and CBAM use an \emph{attention} mechanism in their architectures. `Attention' refers to the notion of directing your focus onto some components of the input or the feature maps via the use of appropriate learnable weights.
In particular, ResNeSt uses a split attention mechanism which breaks the incoming features into sub-groups along the channel dimension, after which attention is applied in each of the sub-groups before merging them together. CBAM uses a Convolutional Block Attention Module to compute feature attention, which contains both a channel and a spatial attention component. 
Network depth, the number of sub-groups considered in ResNeSt 
and the type of attention mechanism are all key factors in determining the representation power of the model. This is also reflected in the results shown in Table \ref{tab:results_2}.

\begin{table}[!htb]
    \centering
    \caption{Results obtained by CBAM and ResNeSt-based backbones on the ESAD test set without using the validation set for training.} 
    \label{tab:results_1}
    \begin{tabular}{lcccc}
    \toprule
         Backbone &$AP_{mean}$ & $AP_{10}$ & $AP_{30}$ & $AP_{50}$  \\
    \midrule
         ResNet-50 & 14.1 & 19.3 & 15.7 & 7.2 \\
         ResNet-101 & 12.9 & 17.7 & 14.3 & 6.4 \\
         CBAM-50 & 13.2 & 18.6 & 14.5 & 6.5 \\
         CBAM-101 & 13.9 & 19.6 & 15.3 & 6.8 \\
         ResNeSt-50 & 13.9 & 18.5 & 15.9 & 7.2 \\
         ResNeSt-101 & \textbf{15.4} & \textbf{20.9} & \textbf{17.3} & \textbf{8.1} \\
         ResNeSt-200 & 0.149 & 0.199 & 0.169 & 0.080 \\
    \bottomrule
    \end{tabular}
\end{table}

\begin{table}[!htb]
    \centering
    \caption{Results obtained by CBAM and ResNeSt-based backbones on the ESAD test set after including the validation set as part of the training data.} 
    \label{tab:results_2}
    \begin{tabular}{lcccc}
    \toprule
         Backbone &$AP_{mean}$ & $AP_{10}$ & $AP_{30}$ & $AP_{50}$  \\
    \midrule
         ResNet-50 & 14.7 & 20.7 & 17.0 & 6.3 \\
         ResNet-101 & 16.0 & 21.7 & 18.8 & 7.5 \\
         CBAM-50 & 13.8 & 20.5 & 15.5 & 5.4 \\
         CBAM-101 & 13.6 & 19.5 & 15.8 & 5.5 \\
         ResNeSt-50 & 15.7 & 22.2 & 17.6 & 7.2 \\
         ResNeSt-101 & 18.3 & 25.2 & \textbf{20.9} & 8.8 \\
         ResNeSt-200 & \textbf{18.9} & \textbf{25.3} & \textbf{20.9} & \textbf{10.2} \\
    \bottomrule
    \end{tabular}
\end{table}

Table \ref{tab:results_2} reports the performance of the above-discussed models when the validation data is added to the training set for training purposes. The numbers show a significant improvement in performance when the training data is so increased. In both Tables \ref{tab:results_1} and \ref{tab:results_2} ResNeSt-based models achieve a much better performance than any other single stage-detector backbones. The more sophisticated attention mechanism of the ResNeSt model can be credited for these significantly higher average precision scores. 

\begin{figure}[ht!]
    \centering
    \includegraphics[width=\textwidth]{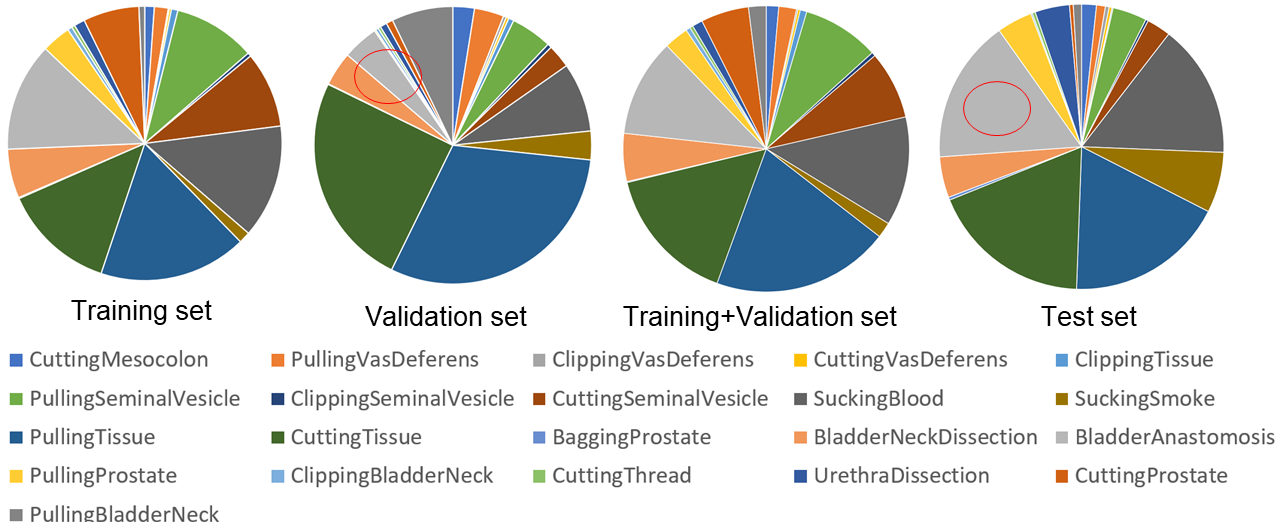}
    \caption{Pie charts showing the percentage of instances for each action class in the training, validation and test sets, plus the combination of training and validation folds.}
    \label{fig:piechart}
\end{figure}

It is interesting to note that ResNeSt-101 performs better than its deeper version ResNeSt-200 (see Table \ref{tab:results_1}). The reason for this lower performance across all IoU values can be traced to its overfitting the training data. As shown in Table \ref{tab:results_2}, ResNeSt-200 does surpass the performance of ResNeSt-101 when the validation data is added to the training set.  Additionally, with this increase in the size of the training set the $AP_{mean}$ for ResNeSt-200 improves from $14.9$ to $18.9$. This reinforces the hypothesis that class imbalance pushes models to give more importance to the most populated classes. 
\\
In Figure \ref{fig:piechart} the distribution of the action classes in each of the splits is represented in the form of a pie chart. It is clear from the pie chart related to the training fold that some classes have an overwhelming number of instance there, while others are very sparsely populated. Merging the training and validation sets not only increases the sample count per class but also balances the overall class distribution. 

The increase in the available training data pushes ResNeSt-200 to learn more robust feature representations when compared to ResNeSt-101. The lower feature representation ability of ResNeSt-101 makes it saturate early due to its shallower nature. Additionally, the inclusion of the validation set in the training set reduces class imbalance, leading in turn to better generalisation on the test set. The higher representation power of ResNeSt-200 also enables it to localise action instances better, a fact which can be observed from the $AP_50$ performance as a higher IoU threshold requires the model to be more accurate at the localisation task. From Table \ref{tab:results_2} it can be seen that ResNeSt-200 achieves $10.2$ average precision, while ResNeSt-101 only achieves an AP of $8.8$.  

\subsection{Influence of data augmentation}

\begin{figure}
    \centering
  \subfloat[Original image]{\includegraphics[width=0.3\textwidth]{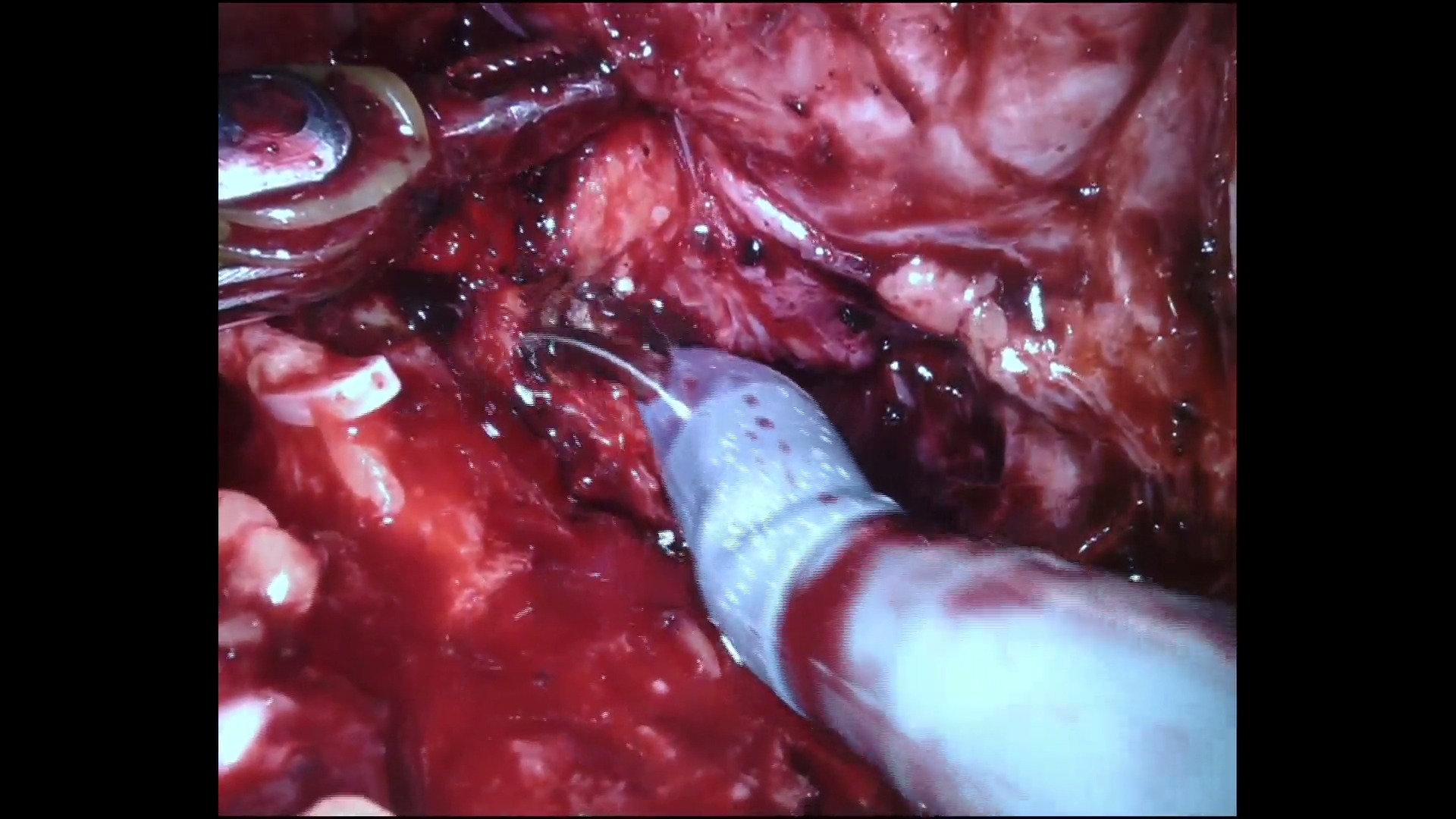}}
  \hfill
  \subfloat[vertical flipping]{\includegraphics[width=0.3\textwidth]{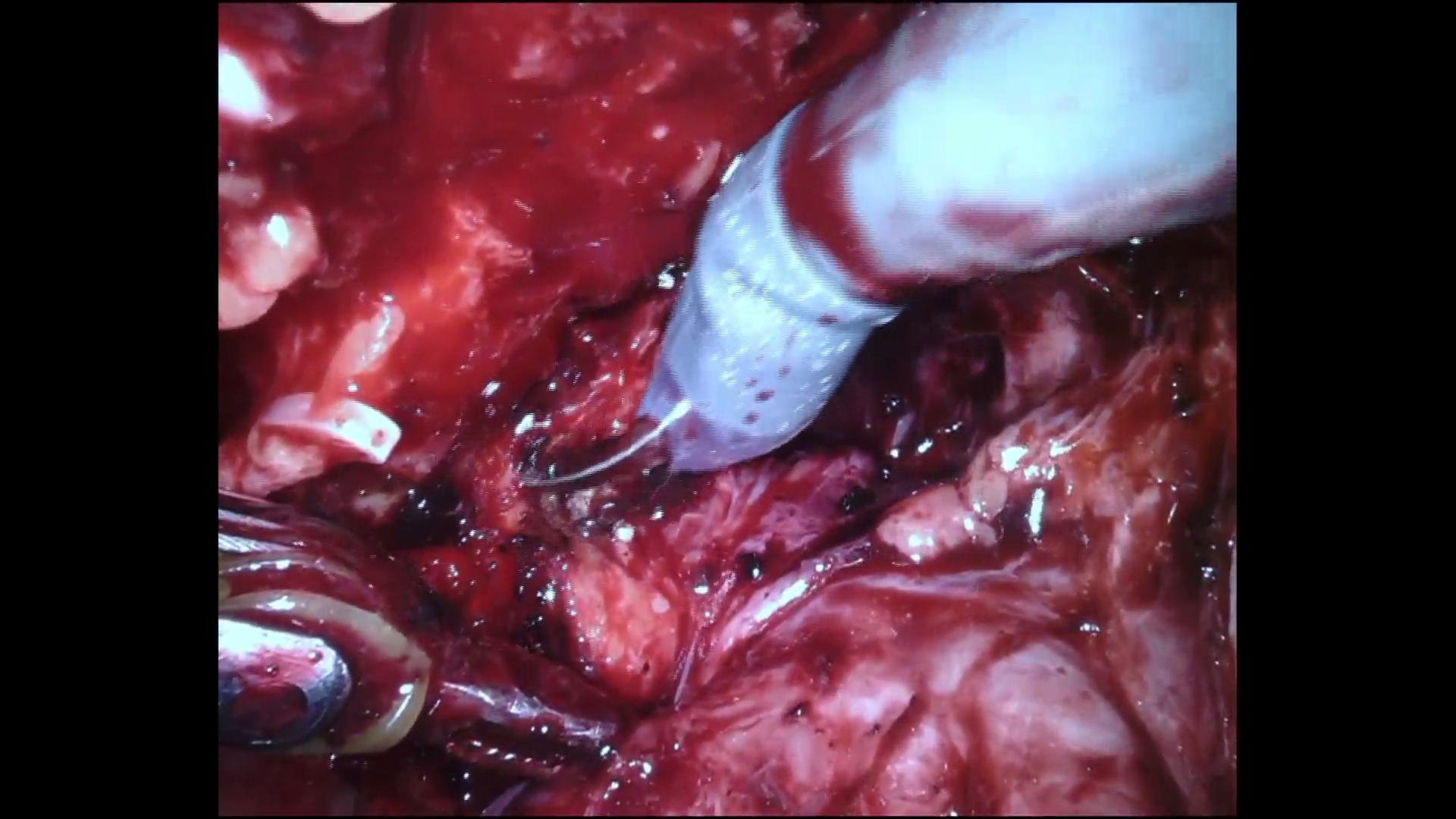}}
  \hfill
  \subfloat[horizontal flipping]{\includegraphics[width=0.3\textwidth]{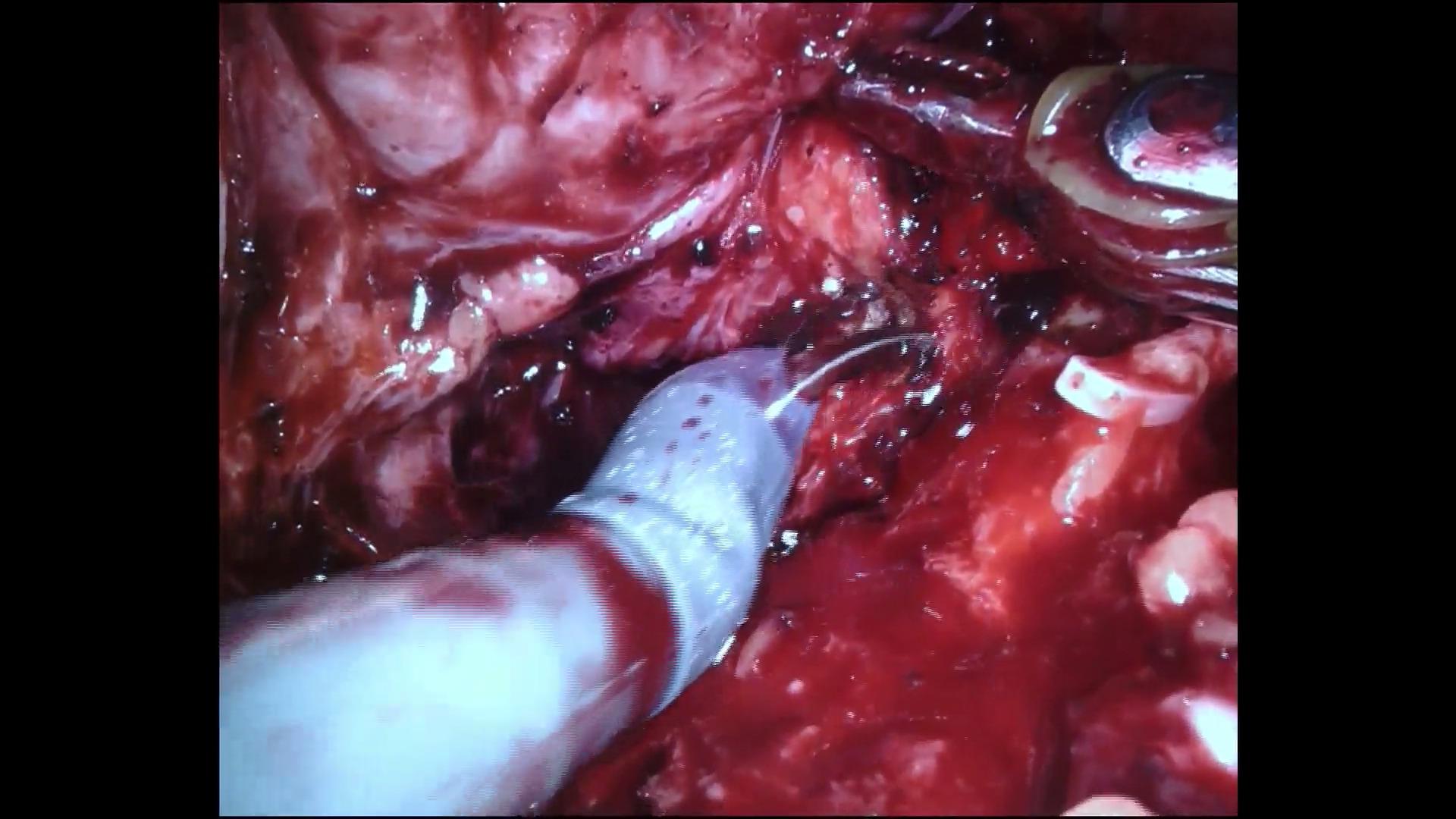}}
  \hfill
  \subfloat[cropping and padding]{\includegraphics[width=0.3\textwidth]{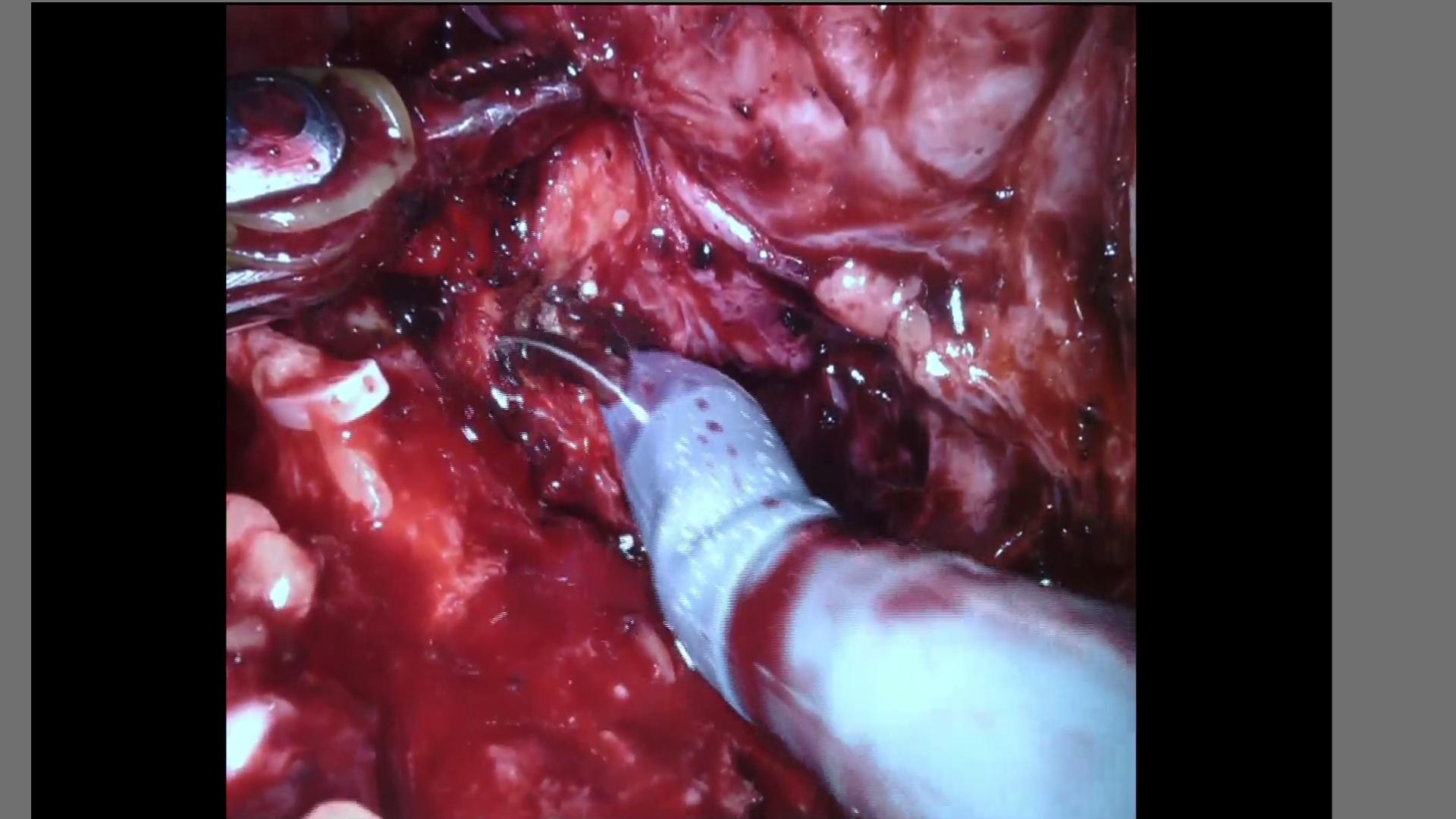}}
  \hfill
  \subfloat[scaling]{\includegraphics[width=0.3\textwidth]{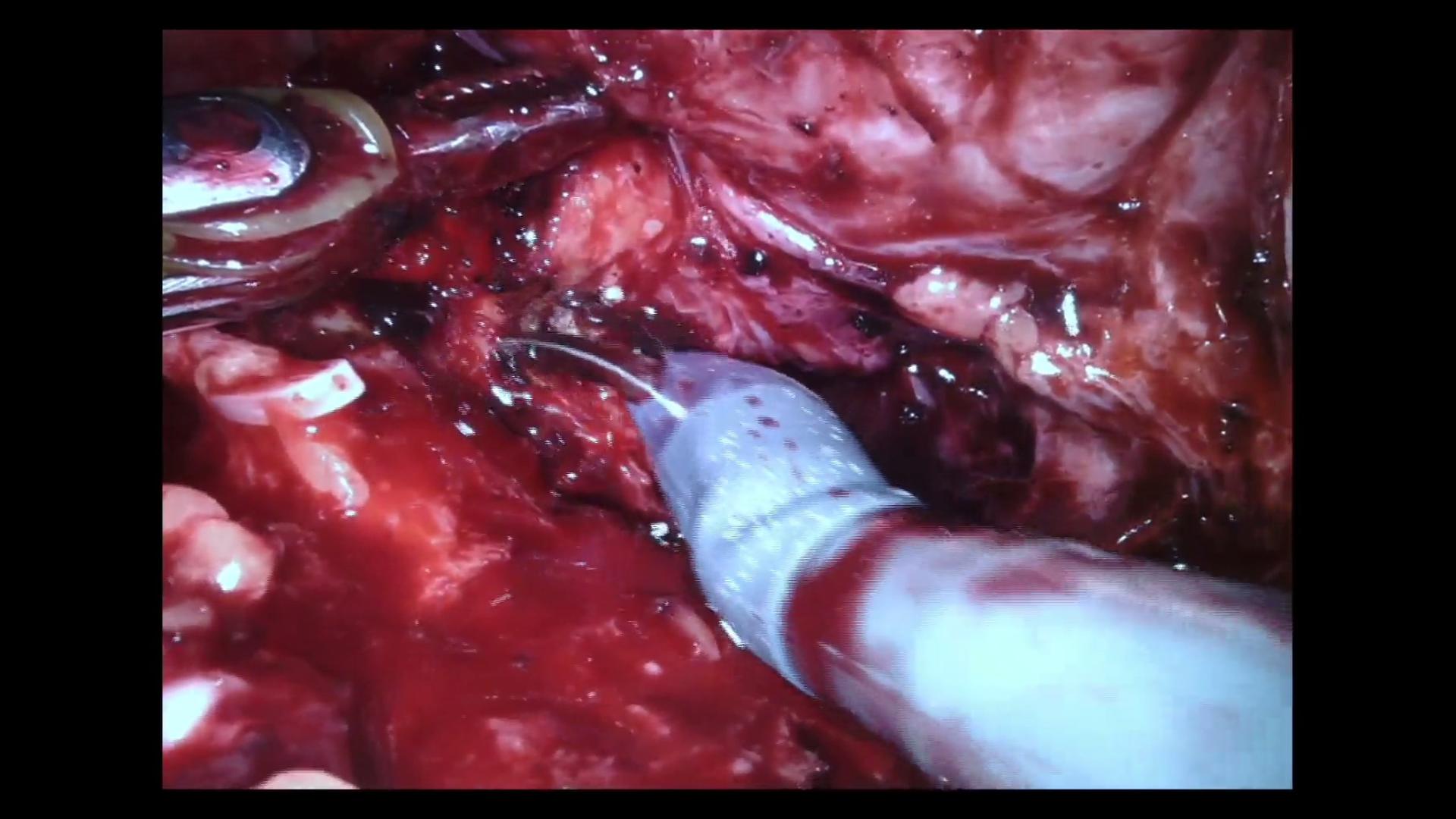}}
  \hfill
  \subfloat[translating]{\includegraphics[width=0.3\textwidth]{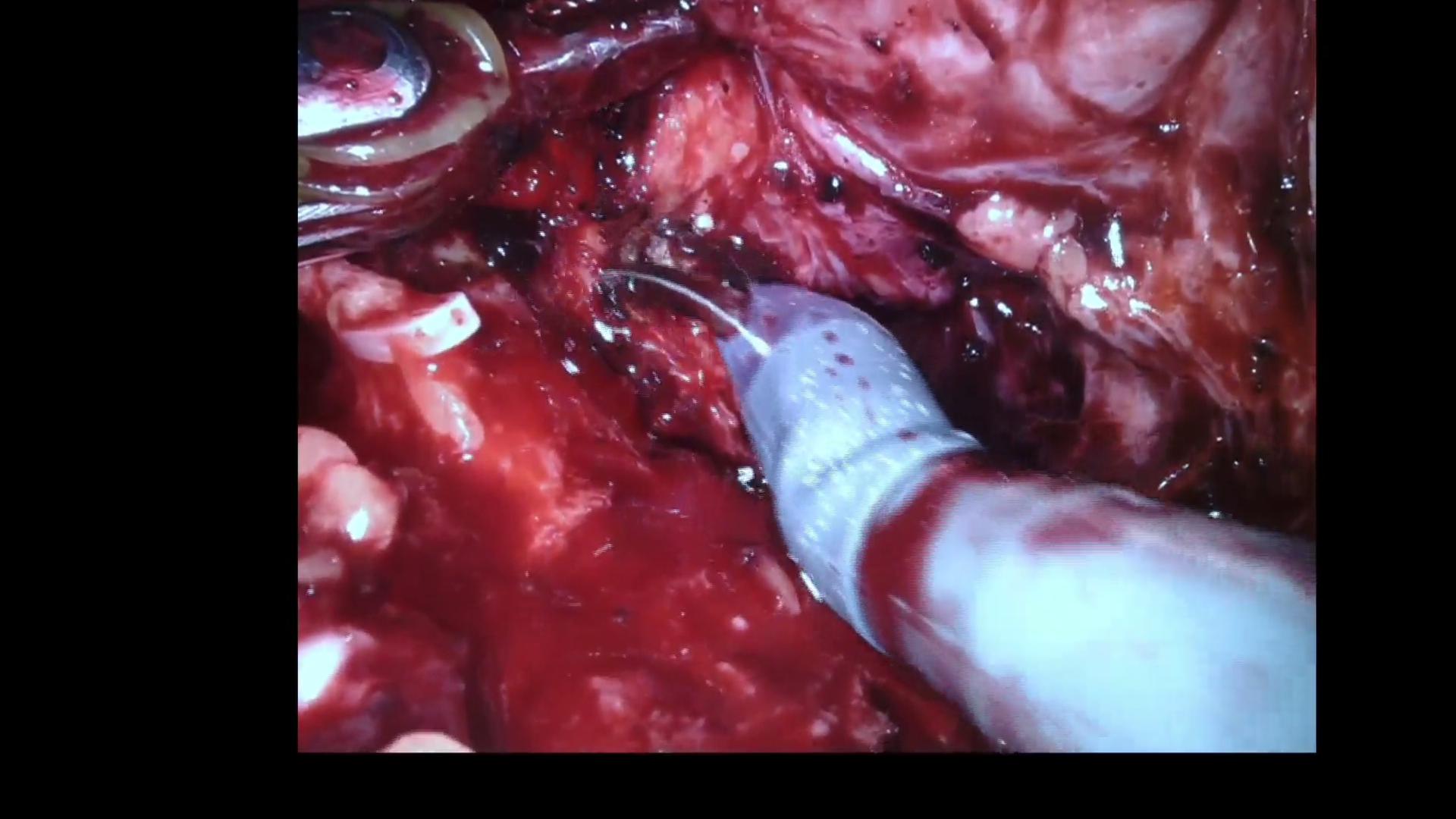}}
  \hfill
  \subfloat[rotating]{\includegraphics[width=0.3\textwidth]{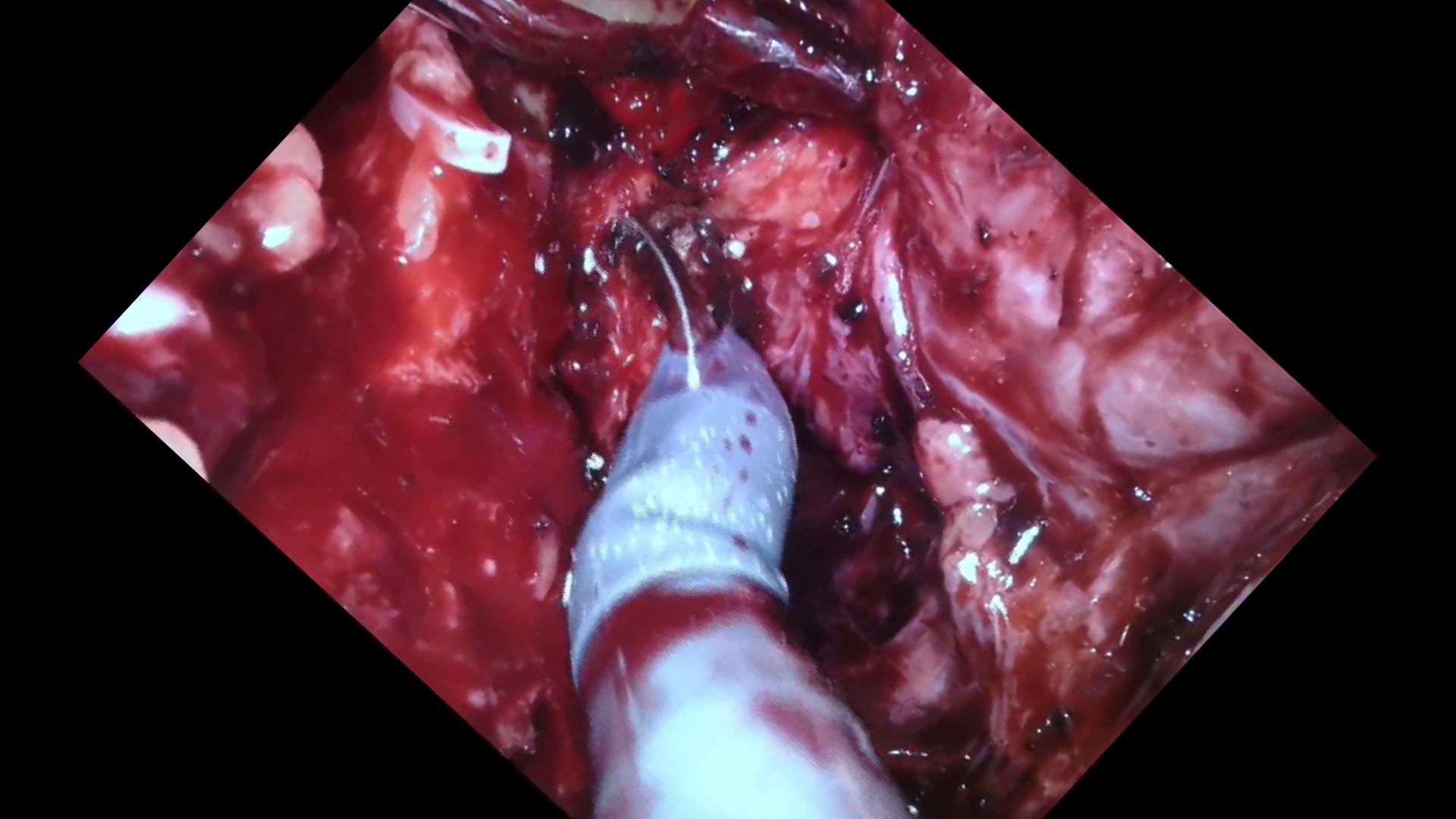}}
  \hfill
  \subfloat[sheering]{\includegraphics[width=0.3\textwidth]{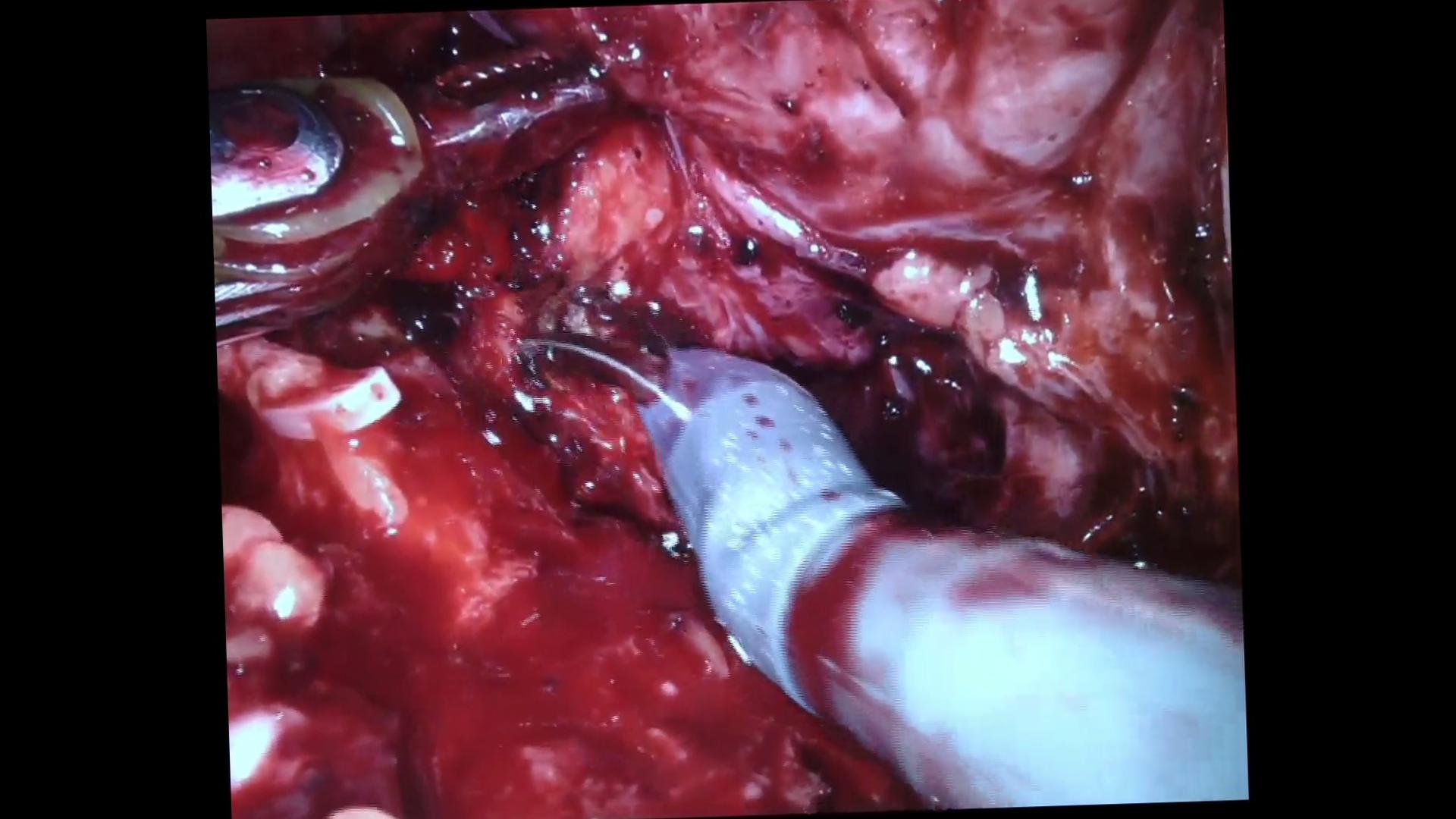}}
  \hfill
  \subfloat[blurring]{\includegraphics[width=0.3\textwidth]{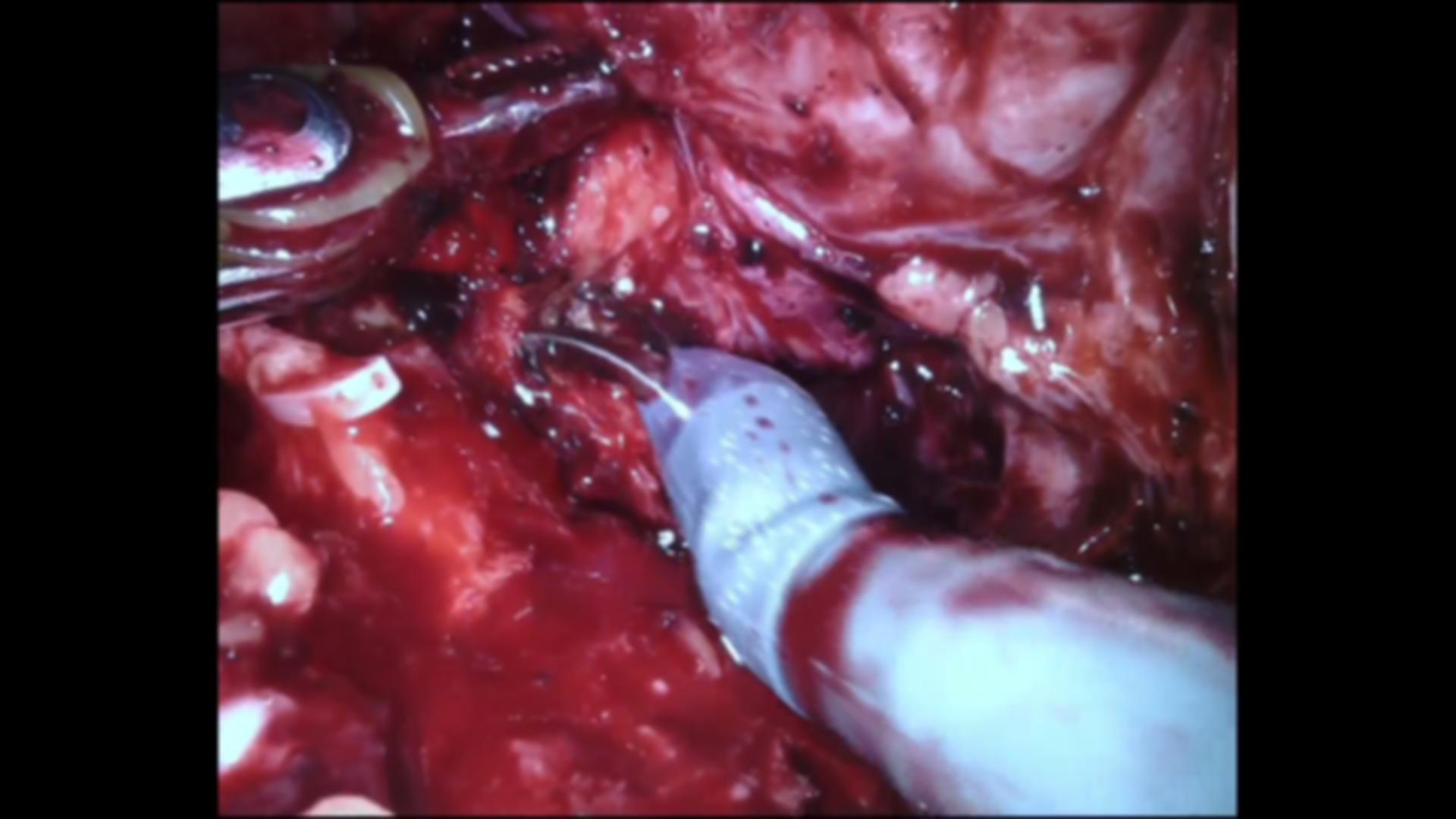}}
  \hfill
  \subfloat[additional noise]{\includegraphics[width=0.3\textwidth]{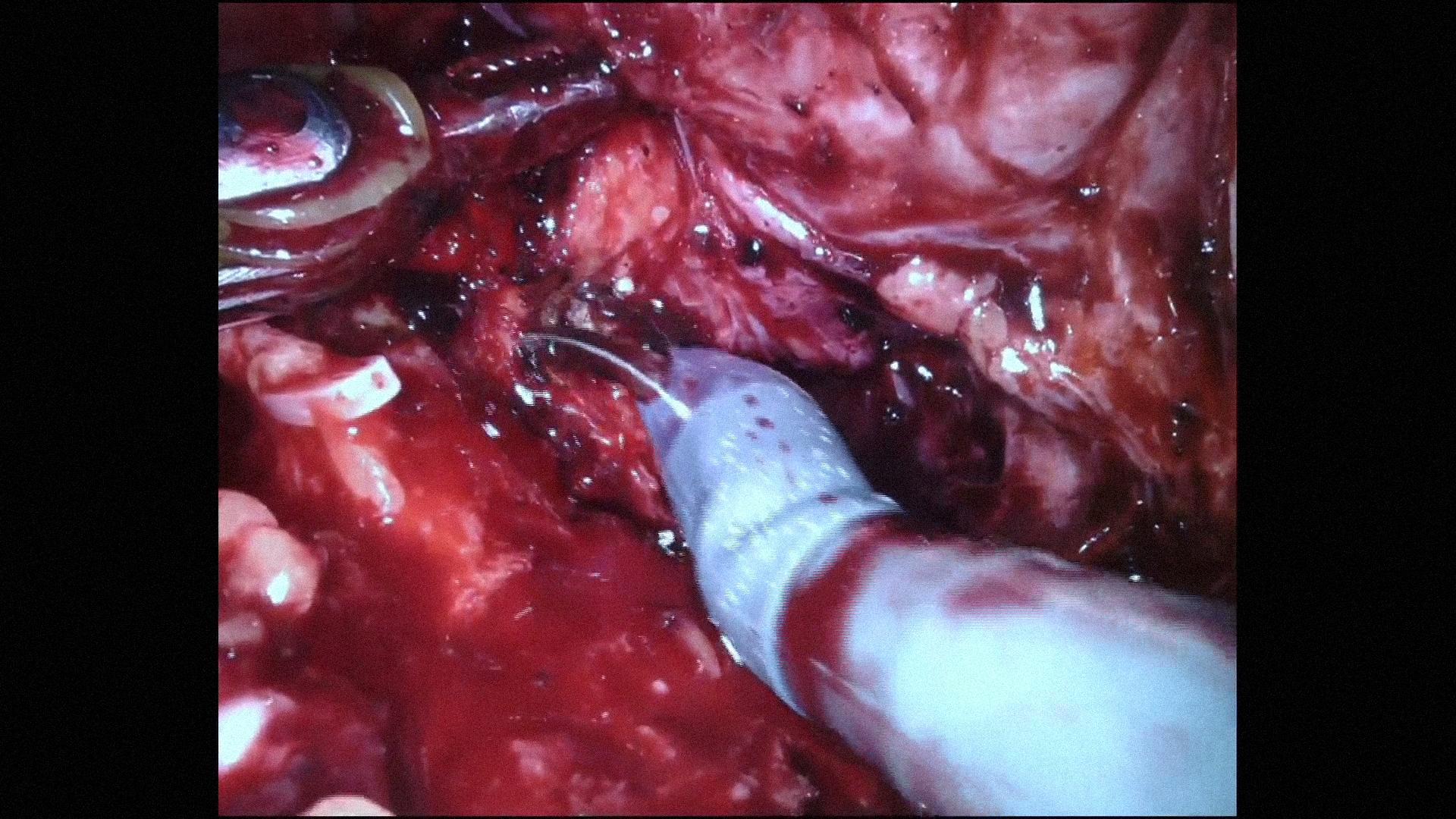}}
  \hfill
  \subfloat[additional frequency noise]{\includegraphics[width=0.3\textwidth]{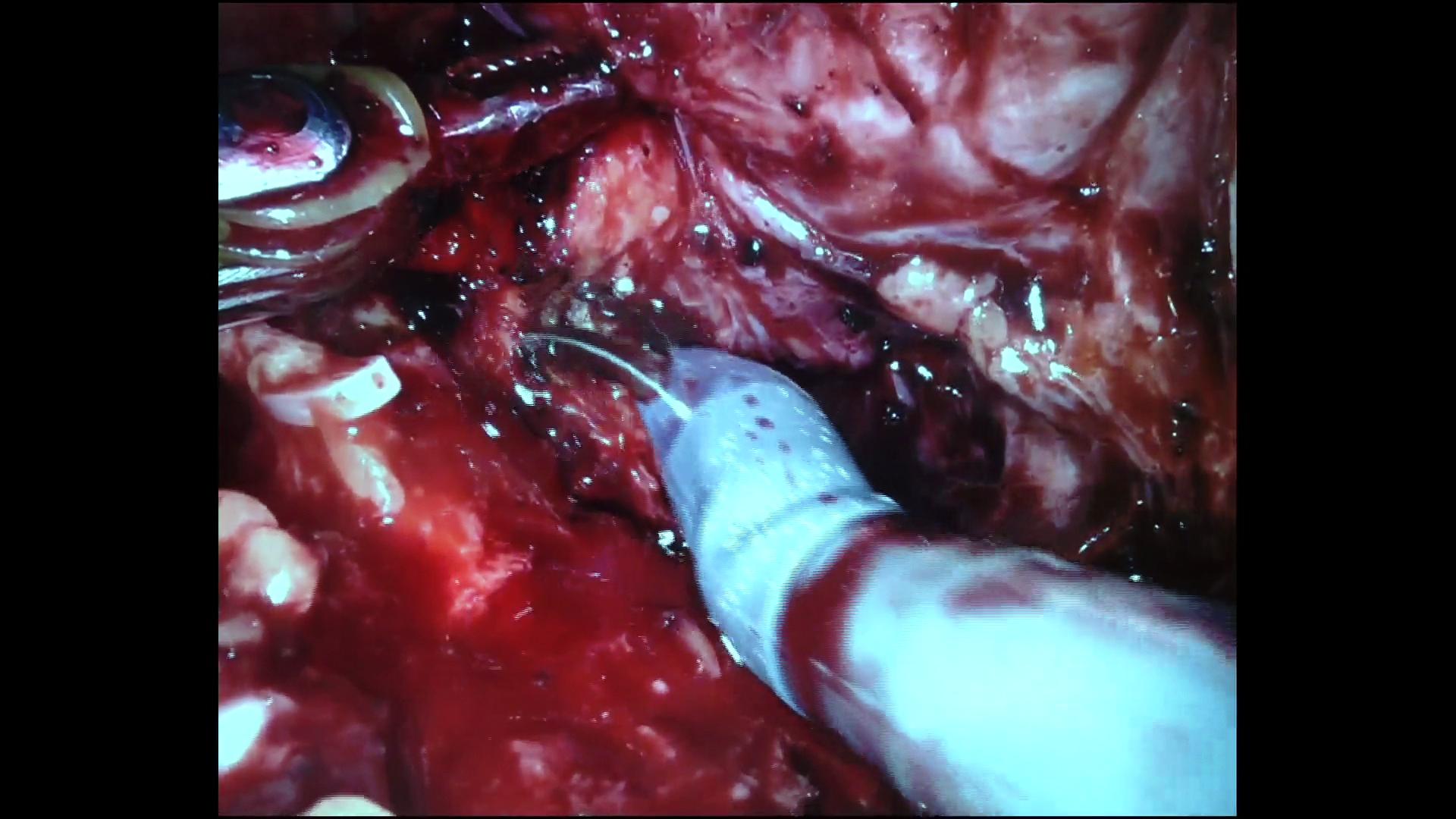}}
  \hfill
  \subfloat[color modification]{\includegraphics[width=0.3\textwidth]{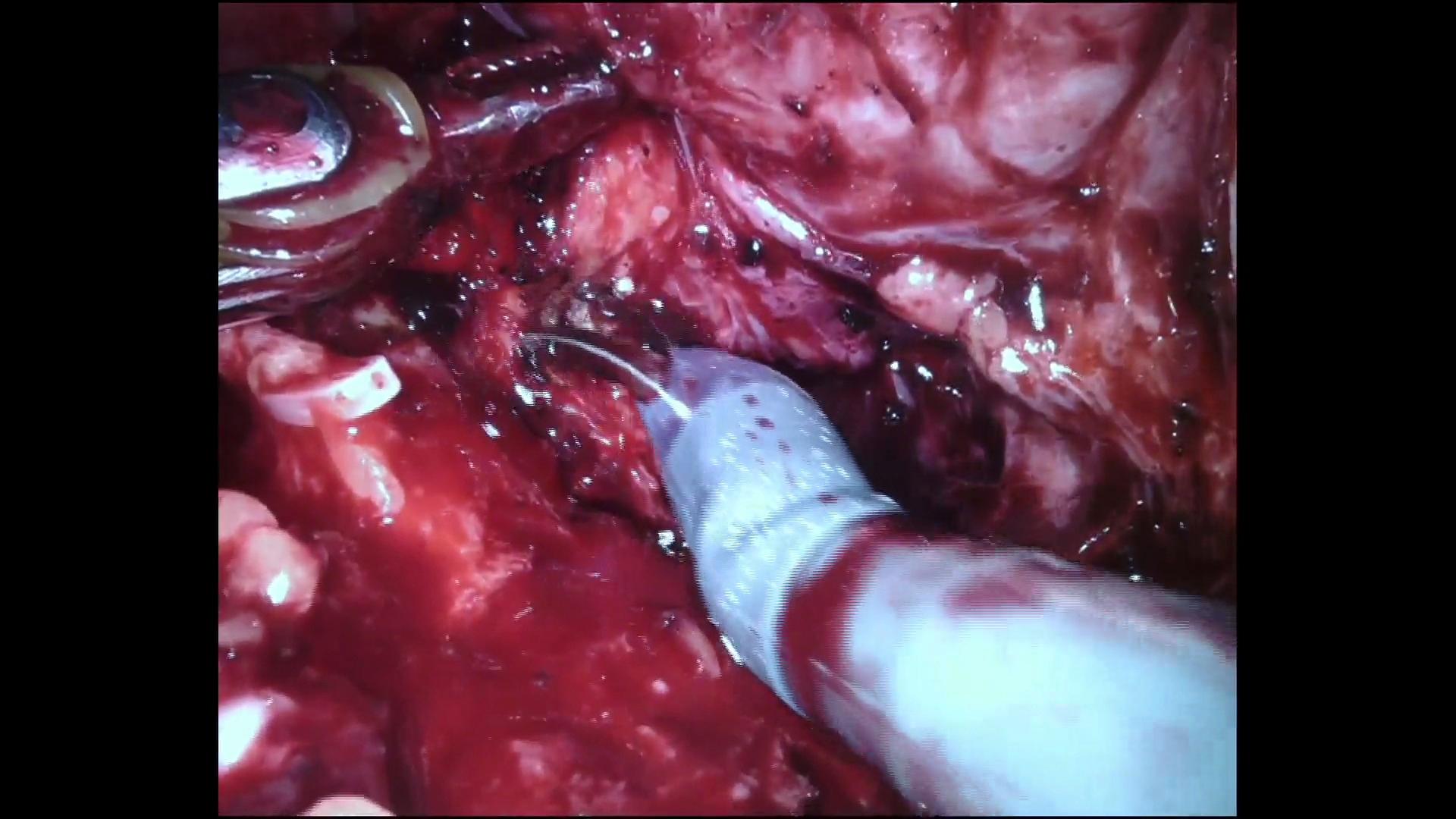}}
  \hfill
  \subfloat[brightness modification]{\includegraphics[width=0.3\textwidth]{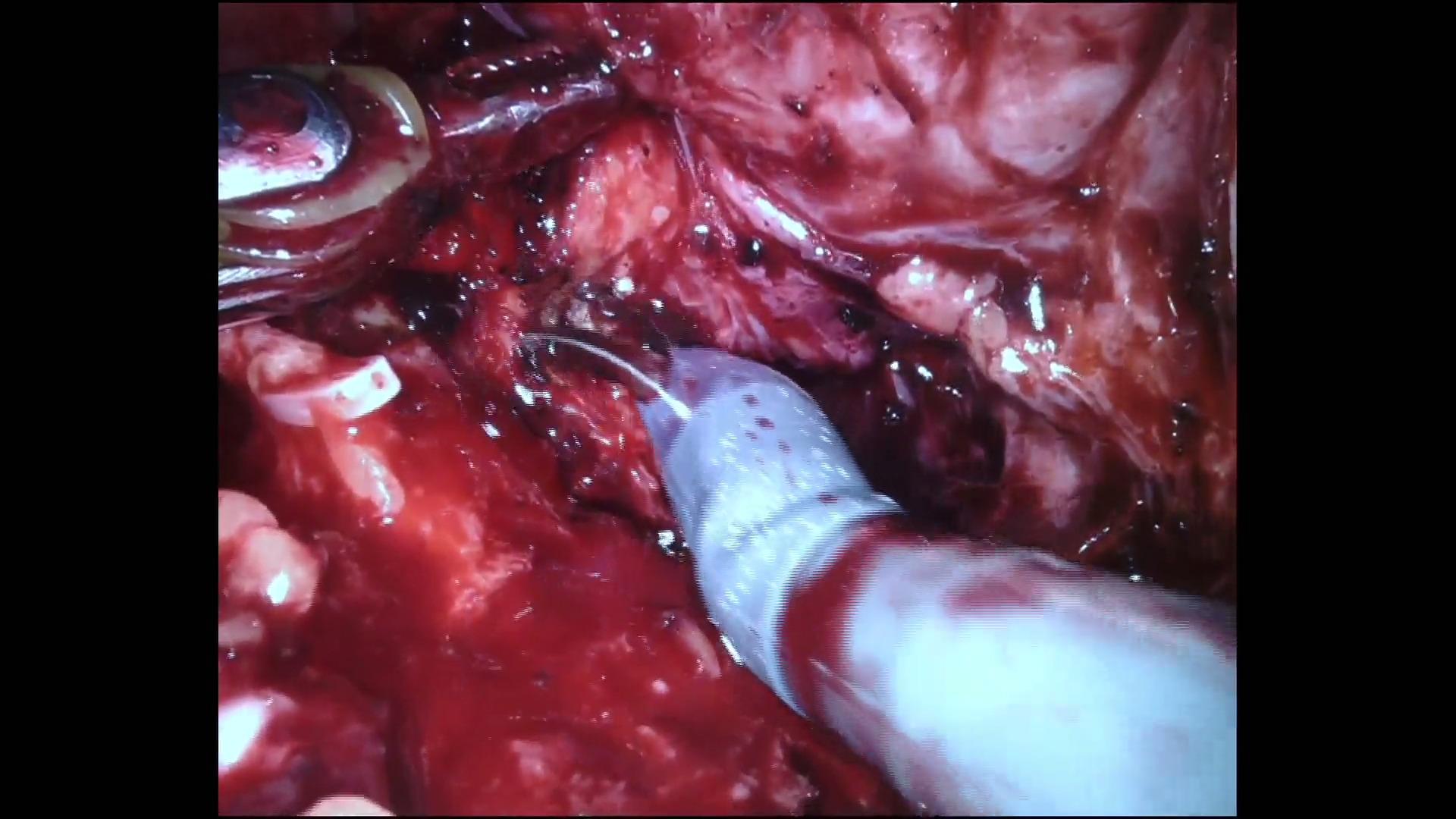}}
  \hfill
  \subfloat[contrast modification]{\includegraphics[width=0.3\textwidth]{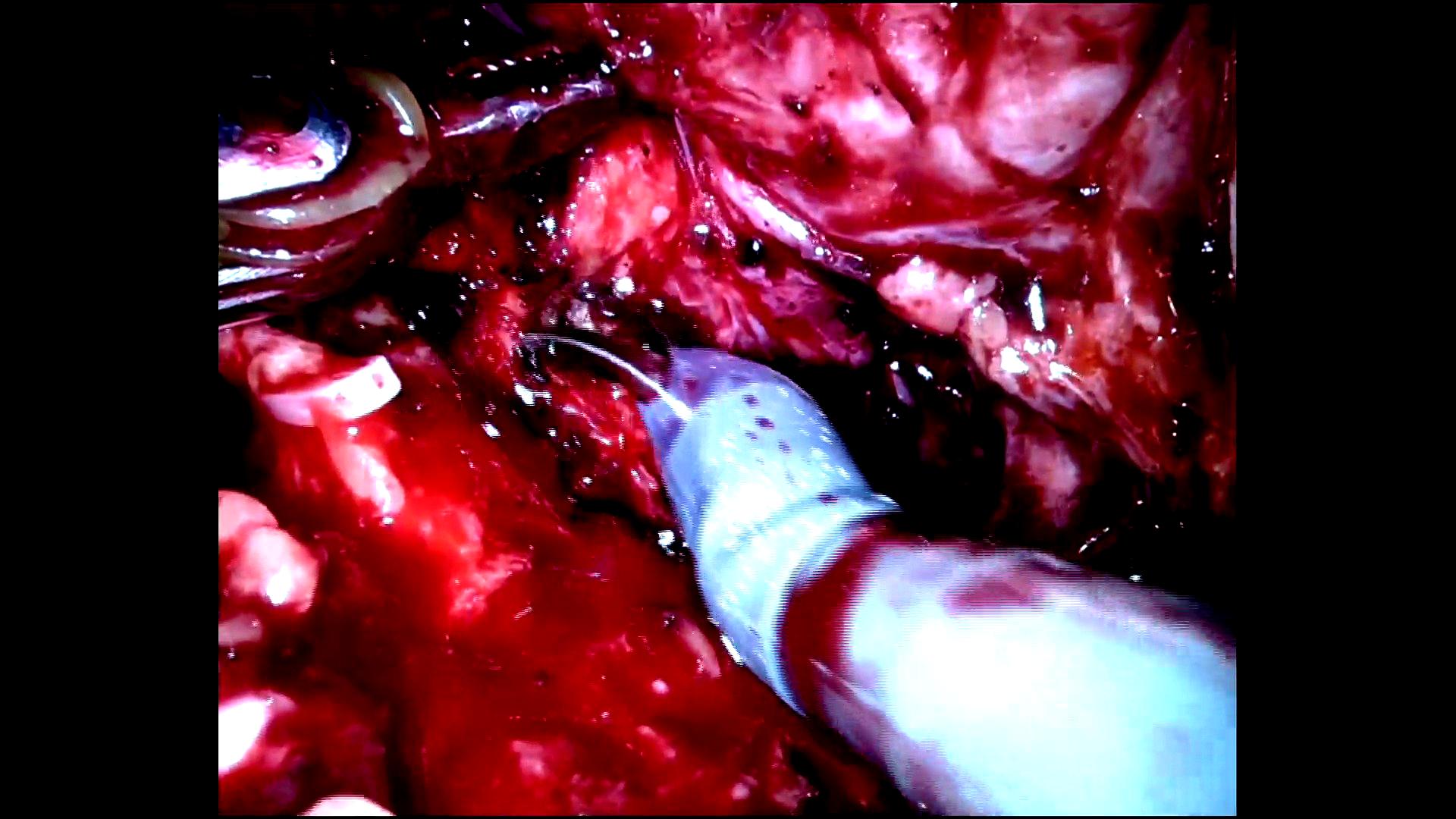}}
  \hfill
  \subfloat[grey scale]{\includegraphics[width=0.3\textwidth]{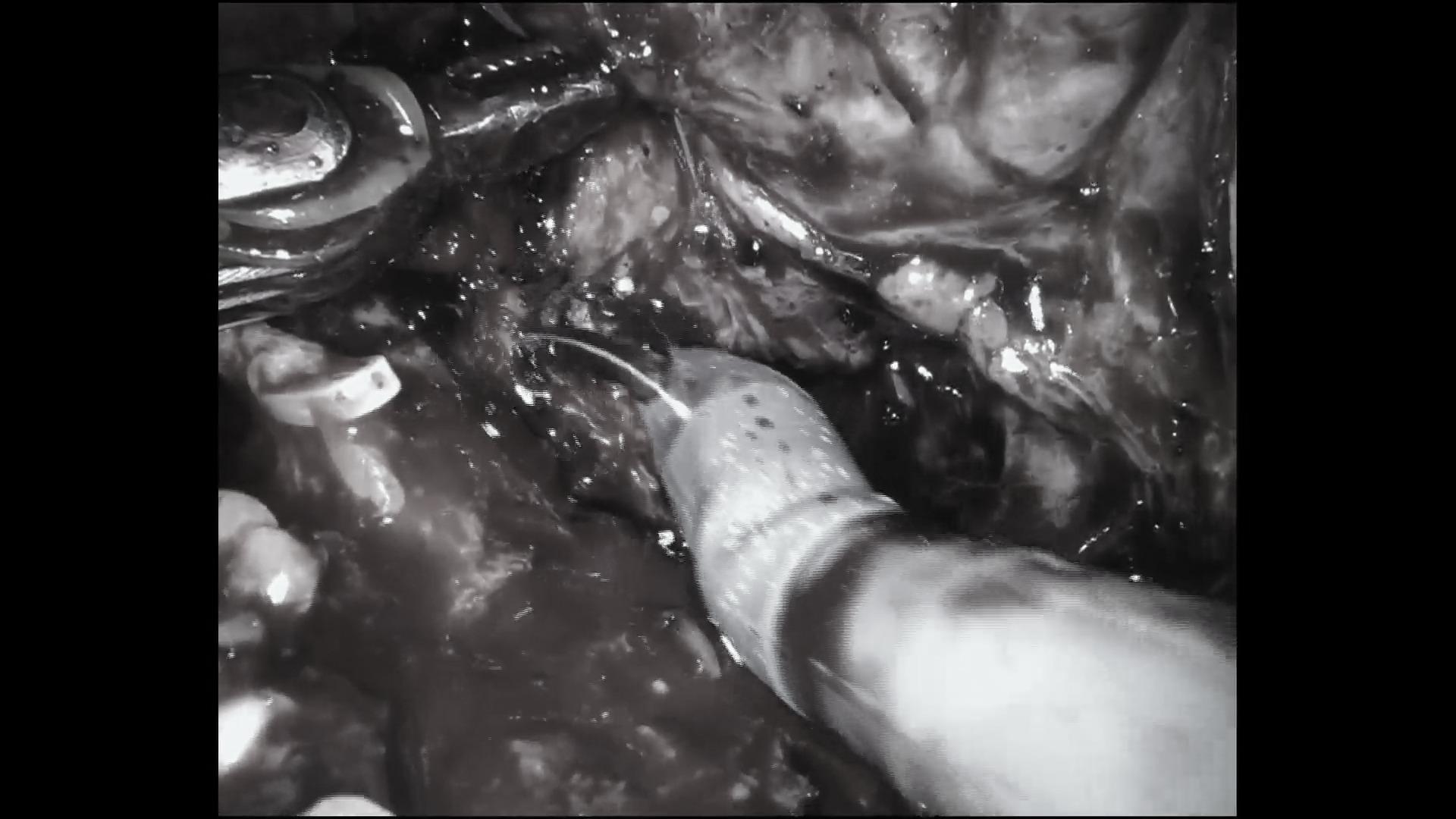}}
  
    \caption{Outputs of different data augmentation methods employed on the ESAD dataset, applied to a sample frame.
    }
    \label{fig:augment}
\end{figure}

One of the top submissions used \emph{data augmentation} methods to boost performance. There were, namely: vertical flipping, horizontal flipping, cropping, padding, scaling, translating, rotating, sheering, blurring, additional noise, additional frequency noise, color modification, brightness modification, contrast modification.
The team maintained that, due to the limitation of the imaging sensors which come with endoscopes, surgical videos are usually of unsatisfactory quality, as they tend to exhibit blur, noise, insufficient brightness or sharpness, and so on. All these issues contribute to make the task even more challenging for computer vision models. Therefore, a strong data augmentation strategy must be set in place. Figure \ref{fig:augment} shows the effects of different augmentation methods on a sample video frame. The figure clearly shows how augmentation can massively increase intra-class variance with the objective of learning more robust models.

\begin{table}[!htb]
    \centering
    \caption{Results from submissions that employ data augmentation in combination with various backbones. The baseline provided by us corresponds to the top row.}
    \label{tab:results_3}
    \begin{tabular}{lcccc}
    \toprule
        Methods & $AP_{mean}$ & $AP_{10}$ & $AP_{30}$ & $AP_{50}$ \\
    \midrule
        ResNet-50 &  16.2 & 21.9 & 17.9 & 8.7 \\
        ResNet-50 + Aug & 17.6 & 23.2 & 19.7 & \textbf{9.9} \\
        SCNet-50 + Aug & 16.4 & 22.9 & 18.5 & 8.0 \\
        ResNeSt-50 + Aug & 17.7 & \textbf{25.0} & \textbf{20.1} & 8.1 \\
        ResNeSt-50 + Aug + LSTM & \textbf{17.9} & \textbf{25.0} & 19.9 & 8.9 \\ 
    \bottomrule
    \end{tabular}
\end{table}

Table \ref{tab:results_3} shows the effects of using the above-mentioned augmentation approaches along with different backbones. 
The first row shows the performance of the baseline model with a ResNet-50 backbone. The use of data augmentation significantly improves the performance of the model even in absence of any additional changes. If the backbone of the model is chosen to be SCNet \cite{liu2020improving} or ResNeSt \cite{zhang2020resnest} (both of which use different types of attention mechanisms), the performance improves even further. SCNet uses a self-calibration mechanism which divides the incoming channels into two sub-branches prior to computing attention. Once again, the ResNeSt-based model achieves the best overall performance on average, whereas ResNet seems to be able to provide more accurate localisation at high IoU. 

\subsection{Exploiting the temporal information}

As the sample images are extracted from the frames of an endoscopic video, one of the top submission set out to exploit the temporal information carried by the video by using several adjacent frames together as the input. 
To obtain a model capable of processing temporal data, a recurrent design was used for both the classification and the localisation heads. 

\begin{figure}[htb]
    \centering
    \subfloat[\label{fig:res31}]{\includegraphics[width=0.8\textwidth]{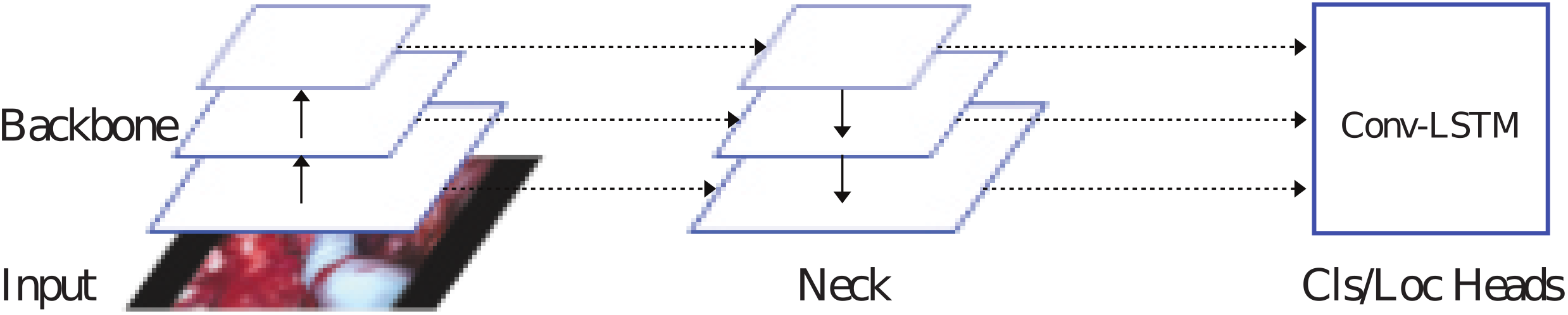}} 
    
    \subfloat[\label{fig:res32}]{\includegraphics[width=0.8\textwidth]{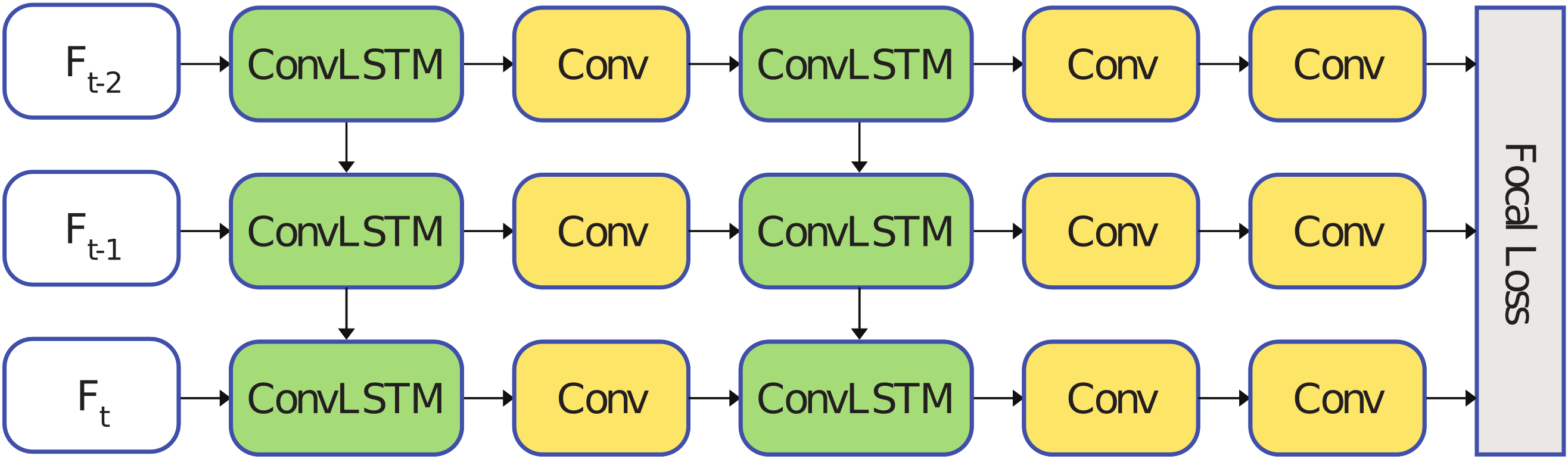}} 
    
    \caption{(a) Architecture of the LSTM-based architecture proposed by one of the top participants. The model exploits the sequential dependence between features from previous frames to predict the action label(s) and bounding box(es) for the current frame. (b) Features from frame $t-2$, $t-1$ and $t$ are alternatively passed through a ConvLSTM layer and a Conv layer. A final convolution layer predicts action labels and bounding boxes. \label{fig:result_3}} 
\end{figure}

As shown in Figure \ref{fig:res31}, three consecutive frames are processed independently through the backbone and the neck module to generate the corresponding spatial features $F_{t-2}$, $F_{t-1}$ and $F_t$. Then, similarly to what done in \cite{li2018recurrent}, a convolutional LSTM (ConvLSTM) module is used in the first and third layers of the heads of RetinaNet. As a consequence, the spatial features extracted by the recursive feature pyramid (RFP) network \cite{qiao2020detectors}, which forms the neck of RetinaNet, are passed to the ConvLSTM module one by one to extract the temporal context (as shown in Figure \ref{fig:res32}). 
In the participants' initial implementation the RFP in RetinaNet was designed for features with 256 channels, but was later reduced to 64 channels due to memory constraint issues. The output of the first ConvLSTM module is further processed by a convolutional layer, another ConvLSTM module and, finally, two extra convolutional layers before the final output. 
The results obtained by this method are shown in the last row of Table \ref{tab:results_3}. There, only the prediction for the last frame of the input triplet of adjacent frames is evaluated.

\subsection{Two-stage detector-based submissions}

We wish to conclude by discussing the submissions proposing a two-stage action detection approach, which all used Faster R-CNN \cite{ren2015faster} as the base model. Various types of backbone alterations along with data augmentation methods were added to improve performance of these models. 

\subsubsection{Data augmentation} \label{sec:faster_aug}

Data augmentation is a key part of training. As we know from the results discussed above, a lower sample count for some classes can be a key driver of a lower mAP. The augmentation methods applied in these challenge entries had two key components: (i) bounding box jitter and (ii) anchor box clustering.

\begin{figure}
    \centering
    \includegraphics[width= \textwidth]{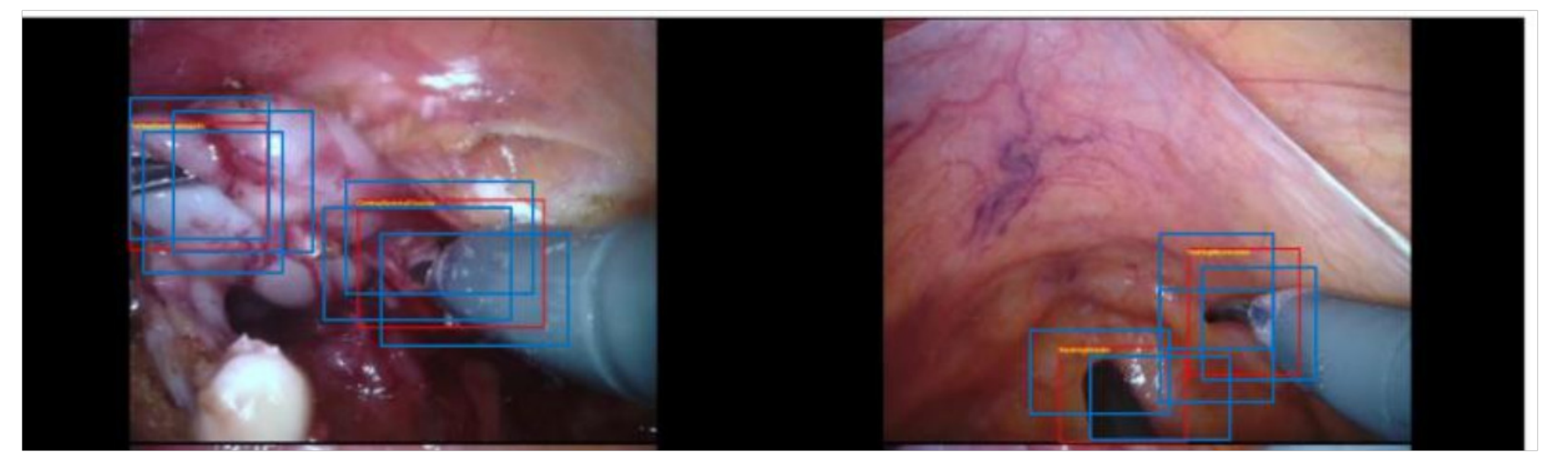}
    \caption{Results of applying random jitter to training bounding boxes. The bounding boxes in red are the original ground truth annotation, while the ones in blue are the those generated by random jitter as in (\ref{eq:jitter}).}
    \label{fig:jitter}
\end{figure}

\emph{Artificial random jitter}. After an analysis of the ESAD dataset, the team found that its endoscope videos are captured from a very close distance. Hence, they are not able to adequately capture the contextual information provided by the surgical scene, unlike what happens in other action detection datasets. In response, they introduced a 'bounding box jitter' technique to improve the ability of the model to capture discriminative features, thus reducing the bias associated with subjective labeling. 
Random artificial jitter is applied as follows:
\begin{equation} \label{eq:jitter}
    Updated = Original * Uniform( [ 0.9, 1.1 ] ).
\end{equation}
'Original' denotes a ground truth bounding box with shape $(x, y, w, h)$, where $x$ and $y$ represent the coordinates of the upper-left corner of the bounding box while $w$ and $h$ are the width and height of the box. 'Uniform' is a uniform probability distribution in the closed interval $[0.9, 1.1]$, whereas 'Updated' is the bounding box so randomly generated. 
The general idea is to use a uniform distribution function to randomly translate, zoom or change the aspect ratio of the original ground-truth bounding boxes so as to increase the diversity of the training data. Examples of the application of jitter are show in Figure \ref{fig:jitter}.

\begin{figure}[htb!]
    \centering
    \includegraphics[width=\textwidth]{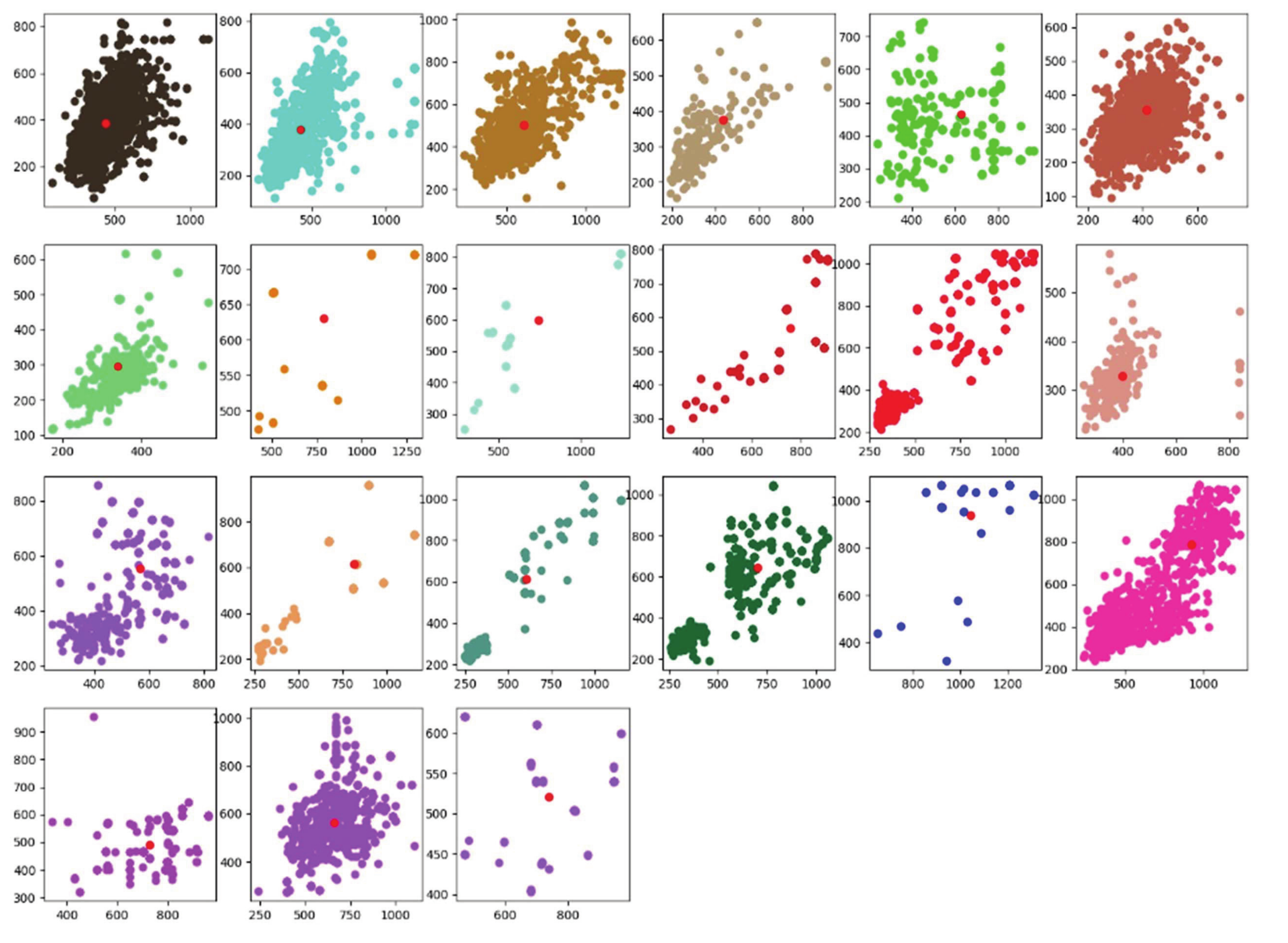}
    \caption{K-means clustering of bounding boxes for each of the 21 action classes in ESAD. Each sub-figure plots the width (x-coordinate) and height (y-coordinate) of the bounding boxes associated with a single class. The red point represents the cluster center.}
    \label{fig:bbox_cluster}
\end{figure}

\emph{Anchor box clustering}. A very important component of the design of detectors is represented by `anchor boxes'. Typically, a fixed set of such anchor boxes is defined with predefined aspect ratio bounds. In order to better fit the needs of the ESAD dataset, the participants used the $K$-means algorithm to cluster the bounding boxes associated with each action class throughout the training data. A class-specific anchor box's aspect ratio for each class was then selected based on the resulting cluster center. Figure \ref{fig:bbox_cluster} shows the resulting bounding box clusters for each class.

As depicted in Figure \ref{fig:bbox_cluster}, the aspect ratio of most cluster centers is situated around 1, showing that the original design ratio for anchors in Faster R-CNN \cite{ren2015faster} (0.5, 1 and 2) is not very applicable here. 
Changing the ratio of these anchors to 1 greatly accelerates the speed of network’s convergence. Firstly, it reduces the complexity of the network and increases the speed of inference. Secondly, because of the box clustering trick, using the anchor initialised by the cluster center to return the other bounding boxes is relatively easy. Finally, it ensure a high proportion of positive samples during the process of allocating positive and negative samples which takes place in Faster R-CNN's Region Proposal Network (RPN) \cite{ren2015faster}.

 
\subsubsection{Model architectures} 

All of the models discussed in this section are based on the Faster R-CNN model \cite{ren2015faster} and results for each of the model are presented in Table \ref{tab:result_5}. 
The first row shows the average precision values obtained with ResNet-101 as the backbone of the Faster R-CNN model. No additional change is made to the model and no data augmentation is used during training. As shown by Table \ref{tab:result_5}, the performance of an out-of the box Faster R-CNN is not very impressive. A significant improvement can be observed simply after employing the data augmentation methods presented in section \ref{sec:faster_aug} prior to training: mAP jumps from $10.97$ to $12.54$, with an even bigger rise in $AP_{10}$. This shows the importance of suitable augmentation in detection tasks.

\begin{table}[!htb]
    \centering
    \caption{Performance on the test set of ESAD dataset achieved by different two-stage approaches based on the Faster-RCNN model.}
    \label{tab:result_5}
    \begin{tabular}{m{5em} m{9em} m{1.5cm} m{1.2cm} m{1.2cm} m{1.2cm}}
    \toprule
        Backbone & Preprocessing & $AP_{mean}$ & $AP_{10}$ & $AP_{30}$ & $AP_{50}$ \\
    \midrule
        ResNet-101 & None & 10.97 & 15.45 & 12.22 & 5.23 \\
        ResNet-101 & Clustering and jitter & 12.54 & 18.11 & 14.15 & 5.37 \\
        ResNet-101 Modified & Clustering and jitter & 16.50 & 23.04 & 18.51 & 7.95 \\
        Ensemble & clustering and jitter & 19.28 & 27.63 & 22.05 & 8.16 \\
    \bottomrule
    \end{tabular}
\end{table}

\begin{figure}
    \centering
    \subfloat{\includegraphics[width=0.33\textwidth]{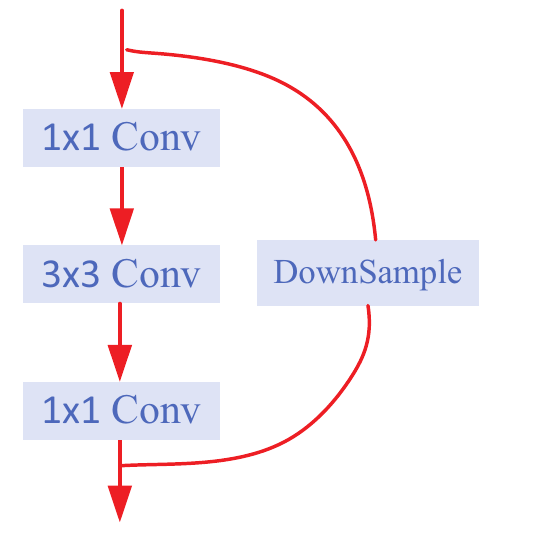}}
    \subfloat{\includegraphics[width=0.33\textwidth]{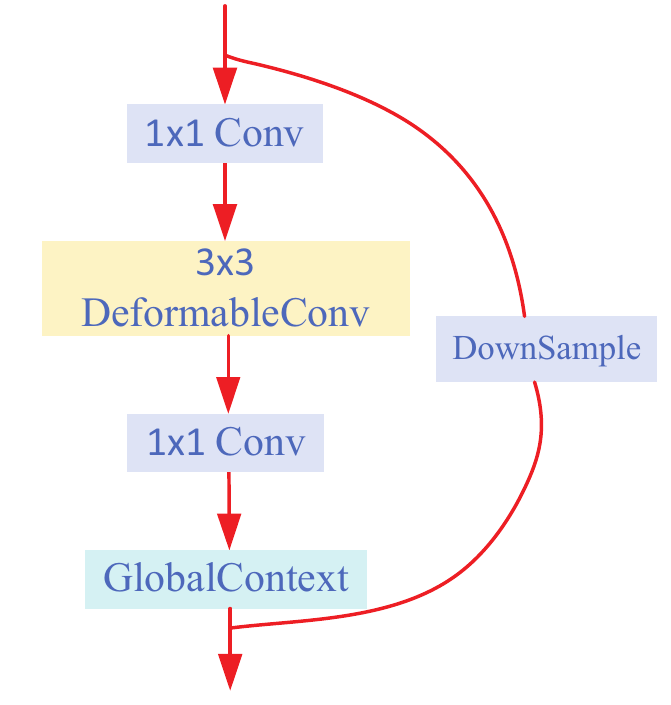}}
    \caption{Block diagram of the original ResNet residual block (left), compared with the proposed modified residual block with deformable convolution (middle) and global context layer (right).}
    \label{fig:resnet_modified}
\end{figure}

\emph{Modified residual block}. In the subsequent model (defined as ResNet-101 Modified in Table \ref{tab:result_5}), a basic building block of ResNet is modified to improve performance. Figure \ref{fig:resnet_modified} shows both the original as well as the modified residual block in ResNet. The original residual block (left) uses a skip connection to connect the input and the output of the residual block. In the participants' modified residual block, a deformable convolution layer \cite{dai2017deformable} (middle) and a global context pooling layer (from GCNet \cite{cao2019gcnet}) are added (right). The deformable convolution kernel adds an offset to the position of each point on the convolution grid. This offset helps decide which point in the neighbourhood of the pixel should be used to perform the convolution operation. In this way, convolution is no longer limited to the traditional regular grid, providing an ability to learn deformable structural features. The global context pooling layer, instead, partitions the feature channels into two streams and computes contextual features from each stream before merging them back together.
\\
Note that ResNet uses four residual blocks. 
In this submission, the modified residual block architecture was used in last three ResNet such layers, rather than all four of them. As shown in Table \ref{tab:result_5}, this amendment further significantly improves model performance, achieving 
an mAP of $16.50$. 

\emph{Ensemble architecture}. The last entry of Table \ref{tab:result_5} relates to an ensemble-based architecture which uses three different feature extractors as part of the backbone, namely ResNeXt-101 \cite{xie2017aggregated}, ResNet-101 \cite{he2016deep} and SENet \cite{hu2018squeeze}. The approach uses weighted boxes fusion \cite{solovyev2019weighted} to merge the bounding boxes predicted by each individual model in the ensemble. The method uses the confidence of the predictions in combination with a predefined IoU threshold to organise and merge the different outputs. In this submission a IoU threshold of 0.5 was used. As the results show, this ensemble architecture is able in achieve the best mAP by far ($19.28$) and was the overall winner of the challenge. 


\section{Conclusion} \label{sec:conclusion}

This paper presents the first-of-its-kind dataset for action detection in surgical videos: SARAS-ESAD.
The dataset was developed by manually annotating real surgical videos collected during prostatectomy (RARP) procedures via an endoscope, and aims to provide a much-needed benchmark for the medical computer vision community to develop and test algorithms for surgical action detection in surgical robotics.

The dataset was used as the basis for a MIDL 2020 challenge. This paper presents the baseline model proposed for the challenge as well as a few selected models derived from the top performing entries.
The baseline model is based on single-shot detector architecture which uses ResNet-based backbones for feature extraction. The model was tested with different backbones and at different input resolutions. For training we used two different loss functions: online hard example mining and focal loss. The focal loss-based model was shown to be able to generalise much better to the challenge's test set. We also found that a larger image size results in better model performance, while average complexity/depth models such as ResNet-34 seem to perform better that `deeper' models. 

A large number of entries were submitted to challenge which used a diverse set of approaches. We have reported on the model architectures and the results of the top submissions. The challenge's results show that data augmentation plays a very important role in model performance. Two-stage detector approaches were able to clearly outperform single-stage ones. The best-performing model, in particular, used an ensemble of feature extractors with different architectures and was able to achieved a mean average precision of $19.28$ despite a large class imbalance.


The results clearly illustrate the level of challenge associated with action detection from surgical videos, as opposed to action performed by physical humans in man-made environment. This shows that much work still needs to be done to achieve the levels of reliability required for action detectors to be usefully and robustly integrated into the autonomous surgical platforms of the future.
\\
In the future we will further investigate the issues here identified, work to conduct more extensive experiments on larger, more significant data and explore the potential of cross-domain approaches able to exploit both data from real procedures as well as videos portraying artificial anatomies (phantoms) to deliver more robust models.

\section{Acknowledgement}

This work was conducted as part of the 'Smart Autonomous Robotic Assistant Surgeon' (SARAS) project, funded by the European Union’s Horizon 2020 research and innovation programme under grant agreement No. 779813.


\bibliographystyle{plainnat}

\begin{thebibliography}{10}
\expandafter\ifx\csname url\endcsname\relax
  \def\url#1{\texttt{#1}}\fi
\expandafter\ifx\csname urlprefix\endcsname\relax\def\urlprefix{URL }\fi
\expandafter\ifx\csname href\endcsname\relax
  \def\href#1#2{#2} \def\path#1{#1}\fi

\bibitem{nepogodiev2019global}
D.~Nepogodiev, J.~Martin, B.~Biccard, A.~Makupe, A.~Bhangu, A.~Ademuyiwa,
  et~al., Global burden of postoperative death, Lancet 393~(401) (2019)
  33139--8.

\bibitem{jhustudy}
M.~Makary, M.~Daniel, Study suggests medical errors now third leading cause of
  death in the u.s.,
  \url{https://www.hopkinsmedicine.org/news/media/releases/study_suggests_medical_errors_now_third_leading_cause_of_death_in_the_us},
  online; accessed 15-April-2020 (2016).

\bibitem{jiang2017artificial}
F.~Jiang, Y.~Jiang, H.~Zhi, Y.~Dong, H.~Li, S.~Ma, Y.~Wang, Q.~Dong, H.~Shen,
  Y.~Wang, Artificial intelligence in healthcare: past, present and future,
  Stroke and vascular neurology 2~(4) (2017) 230--243.

\bibitem{liew2018future}
C.~Liew, The future of radiology augmented with artificial intelligence: a
  strategy for success, European journal of radiology 102 (2018) 152--156.

\bibitem{reddy2019artificial}
S.~Reddy, J.~Fox, M.~P. Purohit, Artificial intelligence-enabled healthcare
  delivery, Journal of the Royal Society of Medicine 112~(1) (2019) 22--28.

\bibitem{hughes2020artificial}
A.~Hughes, et~al., Artificial intelligence-enabled healthcare delivery and
  real-time medical data analytics in monitoring, detection, and prevention of
  covid-19, American Journal of Medical Research 7~(2) (2020) 50--56.

\bibitem{leporini2020technical}
A.~Leporini, E.~Oleari, C.~Landolfo, A.~Sanna, A.~Larcher, G.~Gandaglia,
  N.~Fossati, F.~Muttin, U.~Capitanio, F.~Montorsi, et~al., Technical and
  functional validation of a teleoperated multirobots platform for minimally
  invasive surgery, IEEE Transactions on Medical Robotics and Bionics 2~(2)
  (2020) 148--156.

\bibitem{gu2018ava}
C.~Gu, C.~Sun, D.~A. Ross, C.~Vondrick, C.~Pantofaru, Y.~Li,
  S.~Vijayanarasimhan, G.~Toderici, S.~Ricco, R.~Sukthankar, et~al., Ava: A
  video dataset of spatio-temporally localized atomic visual actions, in:
  Proceedings of the IEEE Conference on Computer Vision and Pattern
  Recognition, 2018, pp. 6047--6056.

\bibitem{weinzaepfel2016human}
P.~Weinzaepfel, X.~Martin, C.~Schmid, Human action localization with sparse
  spatial supervision, arXiv preprint arXiv:1605.05197 (2016).

\bibitem{soomro2012ucf101}
K.~Soomro, A.~R. Zamir, M.~Shah, Ucf101: A dataset of 101 human actions classes
  from videos in the wild (2012).
\newblock \href {http://arxiv.org/abs/1212.0402} {\path{arXiv:1212.0402}}.

\bibitem{sarikaya2020towards}
D.~Sarikaya, P.~Jannin, Towards generalizable surgical activity recognition
  using spatial temporal graph convolutional networks, arXiv preprint
  arXiv:2001.03728 (2020).

\bibitem{petlenkov2008application}
E.~Petlenkov, S.~Nomm, J.~Vain, F.~Miyawaki, Application of self organizing
  kohonen map to detection of surgeon motions during endoscopic surgery, in:
  2008 IEEE International Joint Conference on Neural Networks (IEEE World
  Congress on Computational Intelligence), IEEE, 2008, pp. 2806--2811.

\bibitem{voros2008towards}
S.~Voros, G.~D. Hager, Towards “real-time” tool-tissue interaction
  detection in robotically assisted laparoscopy, in: 2008 2nd IEEE RAS \& EMBS
  International Conference on Biomedical Robotics and Biomechatronics, IEEE,
  2008, pp. 562--567.

\bibitem{kocev2014projector}
B.~Kocev, F.~Ritter, L.~Linsen, Projector-based surgeon--computer interaction
  on deformable surfaces, International journal of computer assisted radiology
  and surgery 9~(2) (2014) 301--312.

\bibitem{van2019weakly}
B.~van Amsterdam, H.~Nakawala, E.~De~Momi, D.~Stoyanov, Weakly supervised
  recognition of surgical gestures, in: 2019 International Conference on
  Robotics and Automation (ICRA), IEEE, 2019, pp. 9565--9571.

\bibitem{azari2019using}
D.~P. Azari, Y.~H. Hu, B.~L. Miller, B.~V. Le, R.~G. Radwin, Using surgeon hand
  motions to predict surgical maneuvers, Human factors 61~(8) (2019)
  1326--1339.

\bibitem{li2016subaction}
Y.~LI, J.~OHYA, T.~CHIBA, R.~XU, H.~YAMASHITA, Subaction based early
  recognition of surgeons' hand actions from continuous surgery videos, IIEEJ
  transactions on image electronics and visual computing 4~(2) (2016) 124--135.

\bibitem{derossi2019}
G.~De~Rossi, N.~Piccinelli, F.~Setti, R.~Muradore, F.~Cuzzolin, Surgical action
  recognition with spatiotemporal convolutional neural networks, in:
  Proceedings of the Hamlyn Symposium on Medical Robotics (HSMR19), 2019.

\bibitem{ji20123d}
S.~Ji, W.~Xu, M.~Yang, K.~Yu, 3d convolutional neural networks for human action
  recognition, IEEE transactions on pattern analysis and machine intelligence
  35~(1) (2012) 221--231.

\bibitem{tran2015learning}
D.~Tran, L.~Bourdev, R.~Fergus, L.~Torresani, M.~Paluri, Learning
  spatiotemporal features with 3d convolutional networks, in: Proceedings of
  the IEEE international conference on computer vision, 2015, pp. 4489--4497.

\bibitem{carreira2017quo}
J.~Carreira, A.~Zisserman, Quo vadis, action recognition? a new model and the
  kinetics dataset, in: proceedings of the IEEE Conference on Computer Vision
  and Pattern Recognition, 2017, pp. 6299--6308.

\bibitem{singh2019recurrent}
G.~Singh, F.~Cuzzolin, Recurrent convolutions for causal 3d cnns, in: IEEE
  International Conference on Computer Vision - First International Workshop on
  Holistic Video Understanding, 2019, pp. 0--0.

\bibitem{donahue2015long}
J.~Donahue, L.~Anne~Hendricks, S.~Guadarrama, M.~Rohrbach, S.~Venugopalan,
  K.~Saenko, T.~Darrell, Long-term recurrent convolutional networks for visual
  recognition and description, in: Proceedings of the IEEE conference on
  computer vision and pattern recognition, 2015, pp. 2625--2634.

\bibitem{wang2016temporal}
L.~Wang, Y.~Xiong, Z.~Wang, Y.~Qiao, D.~Lin, X.~Tang, L.~Van~Gool, Temporal
  segment networks: Towards good practices for deep action recognition, in:
  European conference on computer vision, Springer, 2016, pp. 20--36.

\bibitem{ma2016learning}
S.~Ma, L.~Sigal, S.~Sclaroff, Learning activity progression in lstms for
  activity detection and early detection, in: Proceedings of the IEEE
  Conference on Computer Vision and Pattern Recognition, 2016, pp. 1942--1950.

\bibitem{singh2016multi}
B.~Singh, T.~K. Marks, M.~Jones, O.~Tuzel, M.~Shao, A multi-stream
  bi-directional recurrent neural network for fine-grained action detection,
  in: Proceedings of the IEEE conference on computer vision and pattern
  recognition, 2016, pp. 1961--1970.

\bibitem{yeung2016end}
S.~Yeung, O.~Russakovsky, G.~Mori, L.~Fei-Fei, End-to-end learning of action
  detection from frame glimpses in videos, in: Proceedings of the IEEE
  Conference on Computer Vision and Pattern Recognition, 2016, pp. 2678--2687.

\bibitem{saha2016deep}
S.~Saha, G.~Singh, M.~Sapienza, P.~H. Torr, F.~Cuzzolin, Deep learning for
  detecting multiple space-time action tubes in videos, arXiv preprint
  arXiv:1608.01529 (2016).

\bibitem{kalogeiton2017action}
V.~Kalogeiton, P.~Weinzaepfel, V.~Ferrari, C.~Schmid, Action tubelet detector
  for spatio-temporal action localization, in: Proceedings of the IEEE
  International Conference on Computer Vision, 2017, pp. 4405--4413.

\bibitem{peng2016multi}
X.~Peng, C.~Schmid, Multi-region two-stream r-cnn for action detection, in:
  European conference on computer vision, Springer, 2016, pp. 744--759.

\bibitem{singh2017online}
G.~Singh, S.~Saha, M.~Sapienza, P.~Torr, F.~Cuzzolin, Online real-time multiple
  spatiotemporal action localisation and prediction, in: Proceedings of the
  IEEE International Conference on Computer Vision, 2017, pp. 3637--3646.

\bibitem{gkioxari2015contextual}
G.~Gkioxari, R.~Girshick, J.~Malik, Contextual action recognition with r* cnn,
  in: Proceedings of the IEEE international conference on computer vision,
  2015, pp. 1080--1088.

\bibitem{saha2017amtnet}
S.~Saha, G.~Singh, F.~Cuzzolin, Amtnet: Action-micro-tube regression by
  end-to-end trainable deep architecture, in: Proceedings of the IEEE
  International Conference on Computer Vision, 2017, pp. 4414--4423.

\bibitem{hou2017end}
R.~Hou, C.~Chen, M.~Shah, An end-to-end 3d convolutional neural network for
  action detection and segmentation in videos, arXiv preprint arXiv:1712.01111
  (2017).

\bibitem{hou2017tube}
R.~Hou, C.~Chen, M.~Shah, Tube convolutional neural network (t-cnn) for action
  detection in videos, in: Proceedings of the IEEE International Conference on
  Computer Vision, 2017, pp. 5822--5831.

\bibitem{liu2016ssd}
W.~Liu, D.~Anguelov, D.~Erhan, C.~Szegedy, S.~Reed, C.-Y. Fu, A.~C. Berg, Ssd:
  Single shot multibox detector, in: European conference on computer vision,
  Springer, 2016, pp. 21--37.

\bibitem{tian2013spatiotemporal}
Y.~Tian, R.~Sukthankar, M.~Shah, Spatiotemporal deformable part models for
  action detection, in: Proceedings of the IEEE conference on computer vision
  and pattern recognition, 2013, pp. 2642--2649.

\bibitem{felzenszwalb2008discriminatively}
P.~Felzenszwalb, D.~McAllester, D.~Ramanan, A discriminatively trained,
  multiscale, deformable part model, in: 2008 IEEE Conference on Computer
  Vision and Pattern Recognition, IEEE, 2008, pp. 1--8.

\bibitem{ren2015faster}
S.~Ren, K.~He, R.~Girshick, J.~Sun, Faster r-cnn: Towards real-time object
  detection with region proposal networks, in: Advances in neural information
  processing systems, 2015, pp. 91--99.

\bibitem{jain2014action}
M.~Jain, J.~Van~Gemert, H.~J{\'e}gou, P.~Bouthemy, C.~G. Snoek, Action
  localization with tubelets from motion, in: Proceedings of the IEEE
  conference on computer vision and pattern recognition, 2014, pp. 740--747.

\bibitem{uijlings2013selective}
J.~R. Uijlings, K.~E. Van De~Sande, T.~Gevers, A.~W. Smeulders, Selective
  search for object recognition, International journal of computer vision
  104~(2) (2013) 154--171.

\bibitem{li2018recurrent}
D.~Li, Z.~Qiu, Q.~Dai, T.~Yao, T.~Mei, Recurrent tubelet proposal and
  recognition networks for action detection, in: Proceedings of the European
  conference on computer vision (ECCV), 2018, pp. 303--318.

\bibitem{kay2017kinetics}
W.~Kay, J.~Carreira, K.~Simonyan, B.~Zhang, C.~Hillier, S.~Vijayanarasimhan,
  F.~Viola, T.~Green, T.~Back, P.~Natsev, et~al., The kinetics human action
  video dataset, arXiv preprint arXiv:1705.06950 (2017).

\bibitem{monfortmoments}
M.~Monfort, A.~Andonian, B.~Zhou, K.~Ramakrishnan, S.~A. Bargal, T.~Yan,
  L.~Brown, Q.~Fan, D.~Gutfruend, C.~Vondrick, et~al., Moments in time dataset:
  one million videos for event understanding, IEEE Transactions on Pattern
  Analysis and Machine Intelligence (2019) 1--8\href
  {https://doi.org/10.1109/TPAMI.2019.2901464}
  {\path{doi:10.1109/TPAMI.2019.2901464}}.

\bibitem{goyal2017something}
R.~Goyal, S.~E. Kahou, V.~Michalski, J.~Materzyńska, S.~Westphal, H.~Kim,
  V.~Haenel, I.~Fruend, P.~Yianilos, M.~Mueller-Freitag, F.~Hoppe, C.~Thurau,
  I.~Bax, R.~Memisevic, The "something something" video database for learning
  and evaluating visual common sense (2017).
\newblock \href {http://arxiv.org/abs/1706.04261} {\path{arXiv:1706.04261}}.

\bibitem{caba2015activitynet}
F.~Caba~Heilbron, V.~Escorcia, B.~Ghanem, J.~Carlos~Niebles, Activitynet: A
  large-scale video benchmark for human activity understanding, in: Proceedings
  of the IEEE Conference on Computer Vision and Pattern Recognition, 2015, pp.
  961--970.

\bibitem{sigurdsson2018charadesego}
G.~A. Sigurdsson, A.~Gupta, C.~Schmid, A.~Farhadi, K.~Alahari, Charades-ego: A
  large-scale dataset of paired third and first person videos (2018).
\newblock \href {http://arxiv.org/abs/1804.09626} {\path{arXiv:1804.09626}}.

\bibitem{J-HMDB-Jhuang-2013}
H.~Jhuang, J.~Gall, S.~Zuffi, C.~Schmid, M.~J. Black, Towards understanding
  action recognition, in: Proceedings of the IEEE {I}nternational {C}onference
  on {C}omputer {V}ision (ICCV), 2013, pp. 3192--3199.

\bibitem{liris-harl-2012}
C.~{Wolf}, J.~{Mille}, E.~{Lombardi}, O.~{Celiktutan}, M.~{Jiu},
  M.~{Baccouche}, E.~{Dellandréa}, C.-E. {Bichot}, C.~{Garcia}, B.~{Sankur},
  \href{http://liris.cnrs.fr/publis/?id=5498}{{The LIRIS Human activities
  dataset and the ICPR 2012 human activities recognition and localization
  competition}}, Tech. rep., LIRIS UMR 5205 CNRS/INSA de Lyon/Universit\'{e}
  Claude Bernard Lyon 1/Universit\'{e} Lumi\`{e}re Lyon 2/\'{E}cole Centrale de
  Lyon (2012).
\newline\urlprefix\url{http://liris.cnrs.fr/publis/?id=5498}

\bibitem{daly2016weinzaepfel}
P.~Weinzaepfel, X.~Martin, C.~Schmid, Human action localization with sparse
  spatial supervision, arXiv preprint arXiv:1605.05197 (2016).

\bibitem{ava2017gu}
C.~Gu, C.~Sun, S.~Vijayanarasimhan, C.~Pantofaru, D.~A. Ross, G.~Toderici,
  Y.~Li, S.~Ricco, R.~Sukthankar, C.~Schmid, et~al., Ava: A video dataset of
  spatio-temporally localized atomic visual actions, arXiv preprint
  arXiv:1705.08421 (2017).

\bibitem{twinanda2016endonet}
A.~P. Twinanda, S.~Shehata, D.~Mutter, J.~Marescaux, M.~De~Mathelin, N.~Padoy,
  Endonet: a deep architecture for recognition tasks on laparoscopic videos,
  IEEE transactions on medical imaging 36~(1) (2016) 86--97.

\bibitem{stauder2016tum}
R.~Stauder, D.~Ostler, M.~Kranzfelder, S.~Koller, H.~Feu{\ss}ner, N.~Navab, The
  tum lapchole dataset for the m2cai 2016 workflow challenge, arXiv preprint
  arXiv:1610.09278 (2016).

\bibitem{srivastav2018mvor}
V.~Srivastav, T.~Issenhuth, A.~Kadkhodamohammadi, M.~de~Mathelin, A.~Gangi,
  N.~Padoy, Mvor: A multi-view rgb-d operating room dataset for 2d and 3d human
  pose estimation, arXiv preprint arXiv:1808.08180 (2018).

\bibitem{gao2014jhu}
Y.~Gao, S.~S. Vedula, C.~E. Reiley, N.~Ahmidi, B.~Varadarajan, H.~C. Lin,
  L.~Tao, L.~Zappella, B.~B{\'e}jar, D.~D. Yuh, et~al., Jhu-isi gesture and
  skill assessment working set (jigsaws): A surgical activity dataset for human
  motion modeling, in: Miccai workshop: M2cai, Vol.~3, 2014, p.~3.

\bibitem{gkioxari2015finding}
G.~Gkioxari, J.~Malik, Finding action tubes, in: Proceedings of the IEEE
  conference on computer vision and pattern recognition, 2015, pp. 759--768.

\bibitem{lin2017feature}
T.-Y. Lin, P.~Doll{\'a}r, R.~Girshick, K.~He, B.~Hariharan, S.~Belongie,
  Feature pyramid networks for object detection, in: Proceedings of the IEEE
  conference on computer vision and pattern recognition, 2017, pp. 2117--2125.

\bibitem{he2016deep}
K.~He, X.~Zhang, S.~Ren, J.~Sun, Deep residual learning for image recognition,
  in: Proceedings of the IEEE conference on computer vision and pattern
  recognition, 2016, pp. 770--778.

\bibitem{rothe2014non}
R.~Rothe, M.~Guillaumin, L.~Van~Gool, Non-maximum suppression for object
  detection by passing messages between windows, in: Asian conference on
  computer vision, Springer, 2014, pp. 290--306.

\bibitem{lin2017focal}
T.-Y. Lin, P.~Goyal, R.~Girshick, K.~He, P.~Doll{\'a}r, Focal loss for dense
  object detection, in: Proceedings of the IEEE international conference on
  computer vision, 2017, pp. 2980--2988.

\bibitem{redmon2016you}
J.~Redmon, S.~Divvala, R.~Girshick, A.~Farhadi, You only look once: Unified,
  real-time object detection, in: Proceedings of the IEEE conference on
  computer vision and pattern recognition, 2016, pp. 779--788.

\bibitem{girshick2015fast}
R.~Girshick, Fast r-cnn, in: Proceedings of the IEEE international conference
  on computer vision, 2015, pp. 1440--1448.

\bibitem{zhang2020bridging}
S.~Zhang, C.~Chi, Y.~Yao, Z.~Lei, S.~Z. Li, Bridging the gap between
  anchor-based and anchor-free detection via adaptive training sample
  selection, in: Proceedings of the IEEE/CVF Conference on Computer Vision and
  Pattern Recognition, 2020, pp. 9759--9768.

\bibitem{li2018exploring}
Y.~Li, X.~Bian, M.-C. Chang, S.~Lyu, Exploring the vulnerability of single shot
  module in object detectors via imperceptible background patches, arXiv
  preprint arXiv:1809.05966 (2018).

\bibitem{zhao2019object}
Z.-Q. Zhao, P.~Zheng, S.-t. Xu, X.~Wu, Object detection with deep learning: A
  review, IEEE transactions on neural networks and learning systems 30~(11)
  (2019) 3212--3232.

\bibitem{lin2014microsoft}
T.-Y. Lin, M.~Maire, S.~Belongie, J.~Hays, P.~Perona, D.~Ramanan,
  P.~Doll{\'a}r, C.~L. Zitnick, Microsoft coco: Common objects in context, in:
  European conference on computer vision, Springer, 2014, pp. 740--755.

\bibitem{zhang2020resnest}
H.~Zhang, C.~Wu, Z.~Zhang, Y.~Zhu, Z.~Zhang, H.~Lin, Y.~Sun, T.~He, J.~Mueller,
  R.~Manmatha, et~al., Resnest: Split-attention networks, arXiv preprint
  arXiv:2004.08955 (2020).

\bibitem{woo2018cbam}
S.~Woo, J.~Park, J.-Y. Lee, I.~So~Kweon, Cbam: Convolutional block attention
  module, in: Proceedings of the European conference on computer vision (ECCV),
  2018, pp. 3--19.

\bibitem{liu2020improving}
J.-J. Liu, Q.~Hou, M.-M. Cheng, C.~Wang, J.~Feng, Improving convolutional
  networks with self-calibrated convolutions, in: Proceedings of the IEEE/CVF
  Conference on Computer Vision and Pattern Recognition, 2020, pp.
  10096--10105.

\bibitem{qiao2020detectors}
S.~Qiao, L.-C. Chen, A.~Yuille, Detectors: Detecting objects with recursive
  feature pyramid and switchable atrous convolution, arXiv preprint
  arXiv:2006.02334 (2020).

\bibitem{dai2017deformable}
J.~Dai, H.~Qi, Y.~Xiong, Y.~Li, G.~Zhang, H.~Hu, Y.~Wei, Deformable
  convolutional networks, in: Proceedings of the IEEE international conference
  on computer vision, 2017, pp. 764--773.

\bibitem{cao2019gcnet}
Y.~Cao, J.~Xu, S.~Lin, F.~Wei, H.~Hu, Gcnet: Non-local networks meet
  squeeze-excitation networks and beyond, in: Proceedings of the IEEE
  International Conference on Computer Vision Workshops, 2019, pp. 0--0.

\bibitem{xie2017aggregated}
S.~Xie, R.~Girshick, P.~Doll{\'a}r, Z.~Tu, K.~He, Aggregated residual
  transformations for deep neural networks, in: Proceedings of the IEEE
  conference on computer vision and pattern recognition, 2017, pp. 1492--1500.

\bibitem{hu2018squeeze}
J.~Hu, L.~Shen, G.~Sun, Squeeze-and-excitation networks, in: Proceedings of the
  IEEE conference on computer vision and pattern recognition, 2018, pp.
  7132--7141.

\bibitem{solovyev2019weighted}
R.~Solovyev, W.~Wang, T.~Gabruseva, Weighted boxes fusion: ensembling boxes for
  object detection models, arXiv preprint arXiv:1910.13302 (2019).

\end{thebibliography}


\end{document}